\documentclass[10pt,conference,compsocconf]{IEEEtran}
\usepackage{subfigure} 
\usepackage{threeparttable}
\usepackage{booktabs}
\usepackage{amsfonts}
\usepackage{amssymb}
\usepackage{amsmath}
\usepackage{graphicx}
\usepackage{algorithm}
\usepackage{algorithmic}
\usepackage{url}
\graphicspath{{./figures/}}

\def\R{\mathbb{R}}

\def\I{\mathbb{I}}
\def\E{\mathbb{E}}

\def\rSigma{\mathrm{\Sigma}}

\def\rDelta{\mathrm{\Delta}}

\def\({\left(}
\def\){\right)}
\def\[{\left[}
\def\]{\right]}
\def\nn{\nonumber}

\def\<{\texttt{\hspace{-2pt}[}}
\def\>{\texttt{]}}

\begin{document} 

\title{A Collaborative Kalman Filter for Time-Evolving Dyadic Processes}

\author{
\IEEEauthorblockN{San Gultekin\quad John Paisley}
\IEEEauthorblockA{Department of Electrical Engineering, Columbia University\\
Email: \{sg3108, jpaisley\}$@$columbia.edu}
}

\maketitle


\begin{abstract}
We present the collaborative Kalman filter (CKF), a dynamic model for collaborative filtering and related factorization models. Using the matrix factorization approach to collaborative filtering, the CKF accounts for time evolution by modeling each low-dimensional latent embedding as a multidimensional Brownian motion. Each observation is a random variable whose distribution is parameterized by the dot product of the relevant Brownian motions at that moment in time. This is naturally interpreted as a Kalman filter with multiple interacting state space vectors. We also present a method for learning a dynamically evolving drift parameter for each location by modeling it as a geometric Brownian motion. We handle posterior intractability via a mean-field variational approximation, which also preserves tractability for downstream calculations in a manner similar to the Kalman filter. We evaluate the model on several large datasets, providing quantitative evaluation on the 10 million Movielens and 100 million 
Netflix datasets and qualitative evaluation on a set of 39 million stock returns divided across roughly 6,500 companies from the years 1962--2014.
\end{abstract}

\section{Introduction}
Collaborative filtering is a general and effective method for making pairwise predictions based on historical data. It is most frequently associated with recommendation systems, where the model learns user preference by considering all previous user/object interactions. User rating systems are widespread online. For example, Netflix lets users rate movies on a five star rating system, Amazon allows users to rate products and YouTube allows users to ``like'' or ``dislike'' videos on their website. In cases such as these, a major objective is to recommend content to users in an intelligent way. Collaborative filtering addresses this problem by allowing the preferences of other users to inform the recommendations made for a user of interest \cite{Sarwar:2001}. 

Though typically used for predicting user responses, the main idea behind collaborative filtering can be broadened to include other situations where the outcome of an event is predicted using information about the outcomes of all other events in a dyadic setting. For example, in sporting events one might want to predict whether team A will beat team B by using the results of all other games up to that point. Or one might wish to estimate the probability that batter X will get a hit against pitcher Y at a specific moment in time using all previous at-bats for every batter/pitcher combination. In other problem domains, one might wish to predict the stock prices or exchange rates in a way that takes into consideration the patterns of all other measurements up until that point.

An effective approach to collaborative filtering is via matrix factorization \cite{Koren:2009,Salakhutdinov:2007}. These methods embed users and objects into a low-dimensional latent space and make predictions based upon the relevant dot products. Such methods have proved very powerful since they are related to the low-rank matrix completion problem, wherein it is shown that a sparsely sampled low-rank matrix can be exactly recovered in ideal cases, or very closely otherwise \cite{Candes:2009}. In the Bayesian setting, treating missing data is efficiently handled by the joint likelihood function, which allows for simple closed form updates to the latent embeddings \cite{Salakhutdinov:2008}. 

One common assumption in matrix factorization approaches is that the model is ``static,'' meaning that every user or object has a latent location that is fixed in time and the model learns this single point using all data in a ``batch'' setting. Therefore, these models assume, e.g., that a user's taste in movies is unchanging, or a team's sequence of wins and losses is a stationary process. Though performance of such models can be very good, this assumption is clearly untrue and models that account for dynamic information are more appropriate in principle.

In this paper we present a dynamic collaborative filtering model for temporally evolving data for applications including recommendation systems, currency exchange rate tracking, student/question tutoring systems and athletic competition prediction, among others. The model is inspired by the matrix factorization approach to collaborative filtering, where each object has a location in a latent space. However, rather than fixing these locations, we allow them to drift according to a multidimensional Brownian motion \cite{Cinlar:2011}. Since we are motivated by real-time prediction, we only process each observation in the data stream once. Therefore our problem is ideally suited to large data sets. For inference we introduce a variational approximation \cite{Wainwright:2008} since the posterior distribution of model variables given an event is not in closed form; this allows for closed form updates and preserves tractability for future calculations. The model thus learns a time-evolving probability distribution 
on latent {user/object} locations. From the perspective of a single location, the proposed model can be interpreted as a state-space model with dynamics similar to a Kalman filter \cite{Welch:1995}, and so we call our approach the \emph{collaborative Kalman filter} (CKF). 

In addition, for problems such as stock modeling the drift parameter of the latent Brownian motions cannot be considered to be a fixed value because of the changing volatility of the market. We therefore further develop the CKF by modeling the drift parameter of each latent location with a geometric Brownian motion. This approach will require approximations for tractable inference that nevertheless provide good and meaningful results as shown on a large stock dataset.

We organize the paper as follows: In Section \ref{sec.background} we review the basic matrix factorization approach to collaborative filtering and the Kalman filter for time evolving linear modeling. In Section \ref{sec.ckf} we present the collaborative Kalman filtering model. We derive a variational inference algorithm for this model in Section \ref{sec.vbckf} and show experimental results in Section \ref{sec.experiments}.

\section{Background}\label{sec.background}
Before presenting the collaborative Kalman filter (CKF), we review the basic matrix factorization approach to collaborative filtering and the Kalman filter for dynamic state space modeling. These two approaches form the basis for our model, which we present in Section \ref{sec.ckf}.

\subsection{Collaborative filtering with matrix factorization}\label{sec.cf}
Generally speaking, the matrix factorization approach to collaborative filtering models an incomplete $M \times N$ matrix $Z$ as a function of the product of two matrices, a $K\times M$ matrix $U$ and a $K\times N$ matrix $W$ \cite{Koren:2009}. The columns of $U$ (denoted $u_i$) represent the locations of each user in a latent space and the columns of $W$ (denoted $w_j$) represent the locations of each object in the same latent space. In the probabilistic approach to collaborative filtering using matrix factorization, the observed values in $Z$ have a distribution that depends on the dot products of $U$ and $W$, $z_{ij} \sim p(u_i^T w_j)$. For example, when $Z$ is binary, $p(\cdot)$ could be the logistic or probit link functions, when $m$-ary it could be the ordered probit model, and when real-valued it could be a univariate Gaussian. After learning $U$ and $W$ based on the observed data, predictions of $z_{ij}$ for the unseen pair $(i,j)$ can be made using the distribution function $p(u_i^T w_j)$.

To see how matrix factorization can be interpreted as a collaborative filtering technique, consider the case where $z_{ij} \sim N(u_i^Tw_j,\sigma^2)$. Using a maximum likelihood inference approach for $U$ and $W$, the maximum likelihood update of $u_i$ is a least squares vector using the relevant $w_j$ and $z_{ij}$. If we let the set $\Omega_{u_i}$ contain the index values of objects that user $i$ has rated, this update is
\begin{equation}\label{eqn.pmf}
u_i = \left(\textstyle\sum_{j\in\Omega_{u_i}} w_j w_j^T\right)^{-1}\left(\textstyle\sum_{j\in\Omega_{u_i}}z_{ij}w_j\right).
\end{equation}
Therefore, the design matrix for $u_i$ uses the locations of all objects $w_j$ rated by user $i$. The update for each $w_j$ parallels this update for $u_i$ using the relevant user locations, and so the locations of all other users indirectly influence the update for user $i$ in this way.

A large number of collaborative filtering methods can be viewed as variations and developments of this basic model. A potential drawback of such approaches is that they do not use temporal information. A goal in Section \ref{sec.ckf} will be to show how this model can be easily extended to model the temporal evolution of $u_i$ and $w_j$ using Brownian motion. Since the resulting model has dynamics similar to the Kalman filter, we briefly review this method next.

\subsection{The Kalman filter}\label{sec.kalmanfilter}
As mentioned, a drawback of many collaborative filtering techniques such as those that involve updates similar to Equation (\ref{eqn.pmf}) is that no temporal dynamics are being modeled. When viewing the locations of a user $u_i$ or object $w_j$ as being in a state space, the Kalman filter naturally arises as the first place to look for developing a dynamic representation for matrix factorization models. In this section we provide a brief review of the Kalman filter \cite{Welch:1995}, focusing on the equations and using parameterizations that are relevant to our CKF model.

The Kalman filter models a sequence of observed vectors $y_n \in \R^p$ as linear functions of a sequence of latent state vectors $w_n \in \R^d$ with additive noise. These state vectors evolve according to a first-order Markov process, where the current state equals the previous state plus additive noise. Assuming a Gaussian prior distribution on $w_0$, then for $n = 1,\dots,N$ and zero-mean noise, this is written as follows,
\begin{eqnarray}
w_{n+1}\,|\, w_n~~~ &\sim & N(w_n,\alpha I),\\\label{eqn.Kalman2}
y_{n+1}\,|\, w_{n+1} &\sim & N(Aw_{n+1},\sigma^2 I).
\end{eqnarray}
Inference for the state evolution of $w$ proceeds according to the forward algorithm, which includes two steps: ($i$) marginalizing the previous state to obtain a prior distribution on the current state, and ($ii$) calculating the posterior of the current state given an observation.\footnote{Because we are only interested in real-time prediction in this paper, we do not discuss the backward algorithm used for Kalman smoothing.}

Specifically, let the distribution of the current state be $w_n \sim N(\mu_n,\rSigma_n)$. The distribution of the next state $w_{n+1}$ is calculated by marginalizing the current state,
\begin{eqnarray}\label{eqn.marg1}
p(w_{n+1}) &=& \int_{\R^d} p(w_{n+1}|w_n)p(w_n) dw_n\nn\\
		   &=& N(w_{n+1}|\mu_n,\alpha I + \rSigma_n).
\end{eqnarray}
The drift parameter $\alpha$ controls the volatility of the state vectors and is a measure of drift in one unit of time. (In this paper, we present a method for learning a dynamically evolving value for $\alpha$.)

After observing $y_{n+1}$, the posterior of $w_{n+1}$ is
\begin{eqnarray}\label{eqn.kalman_posterior}
p(w_{n+1}|y_{n+1}) & \propto & p(y_{n+1}|w_{n+1})p(w_{n+1})\nn\\
				   & = & N(w_{n+1}|\mu_{n+1},\rSigma_{n+1}),
\end{eqnarray}
where, after defining $B_n = \alpha I + \Sigma_n$,
\begin{eqnarray}
\rSigma_{n+1} & = & (A^TA/\sigma^2 + B_n^{-1})^{-1},\\
\mu_{n+1} & = & \rSigma_{n+1}(A^Ty/\sigma^2 + B_n^{-1}\mu_n).
\end{eqnarray}

Following Equation (\ref{eqn.marg1}), we can use this posterior distribution on $w_{n+1}$ to find the prior for the next state $w_{n+2}$. This also indicates how $\mu_n$ and $\rSigma_n$ were obtained for step $n$, and so only an initial Gaussian distribution on $w_0$ is required to allow for analytical calculations to be made for all $n$.

The extension to continuous time requires a simple modification: For continuous-time Kalman filters, the variance of the drift from $w_n$ to $w_{n+1}$ depends on the time between these two events. Calling this time difference $\rDelta t_n$, we can make the continuous-time extension by replacing $\alpha I$ with $\rDelta t_n \alpha I$. Therefore $w$ is a Brownian motion \cite{Cinlar:2011}.

\subsection{Related work}
A similar model that addressed spatio-temporal information was presented in \cite{Lu:2009}. A fixed-drift version of the model presented here was originally presented at a machine learning workshop \cite{Paisley:2010} and is most closely related to \cite{Sun:2012,Sun:2014}. Because the models share the same name, we discuss some important differences before presenting our version, and also to highlight our contributions:
\begin{itemize}
 \item Our approach does not require Kalman smoothing.  Therefore, we do not store the sequence of state vectors, which allows for our model to scale to massive data sets. The model in \cite{Sun:2012,Sun:2014} is only designed to handle small-scale data problems.
 \item We derive a variational inference algorithm for learning time-evolving distributions on the latent state space vectors akin to the Kalman filter. We also use a geometric Brownian motion for learning a dynamic drift parameter. Our experiments will show that a fully Bayesian approach on the state space improves predictive ability and learning a dynamic drift gives significant information about the underlying structure of the data set.
 \item We derive a probit model for binary and ordinal observations such as movie ratings instead of directly modeling all data types with Gaussians. We will show that this approach gives an improvement in performance when compared with modeling, e.g., discrete star ratings as real-valued Gaussian random variables.
\end{itemize}

\section{Collaborative Kalman Filtering}\label{sec.ckf}

The motivation behind the collaborative Kalman filter (CKF) is to extend the matrix factorization model described in Section \ref{sec.cf} to the dynamic setting. The continuous-time Kalman filtering framework discussed in Section \ref{sec.kalmanfilter} is a natural means for doing this. The CKF models each latent location of a user or object as moving in space according to a Brownian motion. In our case, these correspond to the sets of vectors $\{u_i\}$ and $\{w_j\}$, which are now functions of time. At any given time $t$, the output for dyad $(i,j)$ uses the dot product $\langle u_i\<t\>, w_j\<t\>\rangle$ as a parameter to a distribution as discussed in Section \ref{sec.cf}. 

In the following discussion of the model we focus on one event (e.g., rating, trial, etc.), which can then be easily generalized to account for every event in a way similar to the Kalman filter. We also present the model in the context of $m$-ary prediction using ordered probit regression \cite{Greene:2011}. For this problem, the real line $\R$ is partitioned into $m$ regions with each region denoting a class, such as star rating. Let $\mathcal{I}_k = (l_k,r_k]$ be the partition for class $k$ where $l_k < r_k$, $r_k = l_{k+1}$, $l_k = r_{k-1}$ and $k=1,\dots,m$. The output for pair $(i,j)$ at time $t$, denoted $z_{ij}\<t\>\in \{1,\dots,m\}$, is obtained by drawing from a Gaussian distribution with the mean $\langle u_i\<t\>, w_j\<t\>\rangle$ and some preset variance $\sigma^2$ and finding which partition it falls within. Therefore, the model assumes an order relation between the $m$ classes, for example that a 4 star rating is closer to a 5 star rating than a 1 star rating. (In the binary special case, this 
order relation no longer matters.) We give a more detailed description of the model below.

\emph{Likelihood model:} Let $u_i\<t\> \in \R^d$ represent the latent state vector for object $i$ at time $t$ and similarly for $w_j\<t\> \in \mathbb{R}^d$. Using the partition $\mathcal{I}$ and the probit function, the probability that $z_{ij}\<t\> = k$ is
\begin{equation}\label{eqn.probitlink}
P(z_{ij}\<t\> = k|u_i,w_j) = \int_{\mathcal{I}_k} N\(y|\langle u_i\<t\>, w_j\<t\>\rangle,\sigma^2\)dy.
\end{equation}
For inference we introduce the latent variable $y_{ij}\<t\>$ and expand Equation (\ref{eqn.probitlink}) hierarchically,
\begin{eqnarray}
z_{ij}\<t\>\,|\,y_{ij}\<t\> &\hspace{-2pt} = \hspace{-2pt}& \textstyle \sum_k k\, \I(y_{ij}\<t\> \in \mathcal{I}_k),\\
y_{ij}\<t\>\,|\,u_i,w_j & \hspace{-2pt}\sim\hspace{-2pt} & N(\langle u_i\<t\>, w_j\<t\>\rangle, \sigma^2).\label{eqn.probitlatent}
\end{eqnarray}
We observe that, unlike most collaborative filtering models, there is no assumption that the pair $(i,j)$ only interacts once, which has practical advantages. For example, teams may play each other multiple times, and even in a user rating setting the user may change his or her rating at different points in time, which contains valuable dynamic information. In related problems such as stock modeling, updated values arrive at regular intervals of time.\footnote{As we will see, in this case $y$ is observed and the likelihood is (\ref{eqn.probitlatent}).}

\emph{Prior model:} At time $t$ we are interested in posterior inference for state vectors $u_i\<t\>$ and $w_j\<t\>$. This requires a prior model, which as we've discussed we take to be a latent multidimensional Brownian motion. Let the duration of time since the last event be $\rDelta_{u_i}^{\<t\>}$ for $u_i$ and $\rDelta_{w_j}^{\<t\>}$ for $w_j$. Also, as with the Kalman filter, assume that we have multivariate normal posterior distributions for $u_i\<t-\rDelta_{u_i}^{\<t\>}\>$ and $w_j\<t-\rDelta_{w_j}^{\<t\>}\>$, with $t-\rDelta_{u_i}^{\<t\>}$ being the time of the last observation for $u_i$ and similarly for $w_j$. We write these distributions as
\begin{eqnarray}\label{eqn.prevposteriors}
u_i\<t-\rDelta_{u_i}^{\<t\>}\> & \hspace{-2mm}\sim\hspace{-2mm} &N(\mu'_{u_i}\<t-\rDelta_{u_i}^{\<t\>}\>,\rSigma'_{u_i}\<t-\rDelta_{u_i}^{\<t\>}\>),\quad\quad\nn\\
w_j\<t-\rDelta_{w_j}^{\<t\>}\> & \hspace{-2mm}\sim\hspace{-2mm} &N(\mu'_{w_j}\<t-\rDelta_{w_j}^{\<t\>}\>,\rSigma'_{w_j}\<t-\rDelta_{w_j}^{\<t\>}\>).
\end{eqnarray}
The posterior parameters $\mu'$ and $\Sigma'$ are also dynamically evolving and indexed by time. As is evident, we are assuming that we have independent posteriors of these variables. Though this is not analytically true, our mean-field variational inference algorithm will make this approximation and so we can proceed under this assumption.

By the continuous-time extension of Equation (\ref{eqn.marg1}), we can marginalize $u_i\<t-\rDelta_{u_i}^{\<t\>}\>$ and $w_j\<t-\rDelta_{w_j}^{\<t\>}\>$ in the interval to obtain the prior distributions on $u_i$ and $w_j$ at time $t$,
\begin{eqnarray}\label{eqn.currprior}
u_i\<t\>& \sim &N(\mu_{u_i}\<t\>,\rSigma_{u_i}\<t\>),\nn\\
w_j\<t\>& \sim &N(\mu_{w_j}\<t\>,\rSigma_{w_j}\<t\>),
\end{eqnarray}
where $$\mu_{}\<t\> = \mu'_{}\<t-\rDelta_{}^{\<t\>}\>,\quad\rSigma_{}\<t\> = \rSigma'_{}\<t-\rDelta_{}^{\<t\>}\> + \rDelta_{}^{\<t\>} \alpha I,$$ for the respective $u_i$ and $w_j$. We note that we use an apostrophe to distinguish priors from posteriors. 

We see that we can interpret the CKF as a Kalman filter with multiple interacting state space vectors rather than a known and fixed design matrix $A$ as in Equation (\ref{eqn.Kalman2}).  We also recall from the Kalman filter that $\alpha I$ is the drift covariance in one unit of time, and so the transition from posterior to prior involves the addition of a small value to the diagonal of the posterior covariance matrix. This is what allows for dynamic modeling; when $\alpha=0$, this is simply an online Bayesian algorithm and the posteriors will converge to a delta function at the mean.

\emph{Hyperprior model:} The drift parameter $\alpha$ controls how much the state vectors can move in one unit of time. When $\alpha$ becomes bigger, the state vectors can move greater distances to better fit the variable $y$, also making these vectors more forgetful of previous information. When $\alpha \rightarrow 0$ the model does not forget any previous information and the state vectors simply converge to a point since the posterior distribution becomes more concentrated at the mean. Therefore, $\alpha$ is clearly an important parameter that can impact model performance.

We develop the CKF model by allowing $\alpha$ to dynamically change in time as well. This provides a measure of volatility that will be especially meaningful in models for stock prices, as we will show. For notational convenience, we define the hyperprior for a shared $\alpha$, but observe that extensions to a $u_i$-- and $w_j$--specific $\alpha$ is straightforward (which we discuss in more detail in Section \ref{sec.vbckf}).

We model $\alpha$ as a geometric Brownian motion by defining $\alpha\<t\> = e^{a\<t\>}$ and letting $a\<t\>$ be a Brownian motion. Therefore, as with the state vectors, if $t-\rDelta_{a}^{\<t\>}$ is the last observed time for $a\<t\>$, the distribution of $a\<t\>$ is
\begin{equation}
 a\<t\> \sim N(a\<t-\rDelta_{a}^{\<t\>}\>,c \rDelta_{a}^{\<t\>}).
\end{equation}
Again there is a drift parameter $c$ that plays an important role and requires a good setting, but we observe that defining $\alpha$ to be an exponentiated Brownian motion has an important modeling purpose by allowing for volatility in time, and is not done simply to avoid parameter setting.

When $\alpha$ is a geometric Brownian motion, the transition from posterior to prior for $u_i$ and $w_j$ needs to be modified. In Equation (\ref{eqn.currprior}) the constant value of $\alpha$ allowed for $\Sigma\<t\>$ to be calculated by adding a time-scaled $\alpha I$ to the previous posterior. In this case, the update is modified for, e.g., $u$ by performing the integration implied in Equation (\ref{eqn.currprior}),
\begin{equation}\label{eqn.stoch_int}
 \rSigma_{u}\<t\> = \rSigma'_{u}\<t-\rDelta_{u}^{\<t\>}\> + I \textstyle\int_{t-\rDelta_{u}^{\<t\>}}^t e^{a\<s\>} ds.
\end{equation}
We will derive a simple approximation of this this stochastic integral for inference.

\section{Variational Inference for the CKF}\label{sec.vbckf}
Having defined the model prior, we next turn to posterior inference. Since the CKF model is dynamic, we treat posterior inference in the same way by learning a time-evolving posterior distribution on the underlying vectors $\{u_i\}$ and $\{w_j\}$ (and the hidden data $y_{ij}\<t\>$ when the observations are from a latent probit model). We also learn the Brownian motion controlling the drift of each latent state vector, $a\<t\>$, using a point estimate. We therefore break the presentation of our inference algorithm into two parts: One dealing with $u$, $w$ and $y$, and one dealing with $a$. We present an overview of the inference algorithm in Algorithm \ref{alg.ckf}.

\subsection{An approximation for the hyperprior model}
Before discussing our inference algorithm, we first present an approximation to the stochastic integral in Equation (\ref{eqn.stoch_int}) that will lead to efficient updates of the Brownian motion $a\<t\>$. To this end, we first introduce the following approximation,
\begin{equation}
 \textstyle\int_{t-\rDelta^{\<t\>}}^t e^{a\<s\>} ds ~ \approx ~ e^{a\<t\>}\Delta^{\<t\>}.
\end{equation}
We recall from the description of the model that $a\<t\>$ can be shared or vector-specific. That is, e.g., $a\<t\>$ could be unique for each $u_i$, or shared among $\{u_i\}$ (in which case an appropriate subscript notation for $a$ could be added). From the perspective of a generic multidimensional Brownian motion, the generative structure for this approximation is
\begin{eqnarray}
 u\<t\> &\sim& N(\mu_u'\<t-\Delta_u\>,\Sigma_u'\<t-\Delta_u\> + e^{a\<t\>}\Delta_u I),\nn\\
a\<t\>&\sim& N(a\<t-\Delta_a\>,c\Delta_a).
\end{eqnarray}
That is, with this approximation we first draw the log drift value at time $t$, $a\<t\>$, according to its underlying Brownian motion. We then approximate the drift parameter as being constant in the unobserved interval. Clearly, as the intervals between observations become smaller, this approximation becomes better. For inference we will learn a point estimate of the Brownian motion $a\<t\>$.

\begin{algorithm}[t]
\caption{CKF parameter updates at time $t$}\label{alg.ckf}
\begin{algorithmic}[1]
\STATE To update the parameters of dyad $(i,j)$ at time $t$:\vspace{1pt}
\FOR{$iteration=1,\dots, T$}\vspace{1pt}
\STATE Update $q(y_{ij}\<t\>)$ as in Equation (\ref{eqn.m}).\vspace{1pt}
\STATE Update $q(u_i\<t\>)$ as in (\ref{eqn.muUp}).\vspace{1pt}
\STATE Update $q(w_j\<t\>)$ by modifying (\ref{eqn.muUp}) as noted in text.\vspace{1pt}
\STATE Update $a_{u_i}\<t\>$ as in (\ref{eqn.a}).\vspace{1pt}
\STATE Update $a_{w_i}\<t\>$ by modifying (\ref{eqn.a}) as noted in text.\vspace{1pt}
\ENDFOR
\STATE Notes: The following changes can be made. \\
$\bullet$ When $y$ is observed, skip Step 3 and directly use $y$ in 

~~\, Steps 4 and 5.\\
$\bullet$ When modeling a shared $a_u$ among all $\{u\}$, the update

~~\, is changed as noted in text, and similarly for $w$.
\end{algorithmic}
\end{algorithm}

\subsection{A variational approximation}
Focusing on the latent state-space variables first, the posterior of an event at time $t$ is
\begin{equation}\label{eqn.jointposterior}
p(u_i\<t\>,w_j\<t\>,y_{ij}\<t\>|z_{ij}\<t\>,a_{u_i},a_{w_j}) \propto  \hspace{.5in}
\end{equation}
\begin{equation}\nonumber
~~~~p(z_{ij}\<t\>|y_{ij})p(y_{ij}\<t\>|u_i,w_j)p(u_i\<t\>|a_{u_i})p(w_j\<t\>|a_{w_j})
\end{equation}
Unlike the Kalman filter the full posterior of the CKF is not analytically tractable. Typically when this occurs with nonlinear Kalman filters sampling methods such as sequential Monte Carlo are employed as an approximation \cite{Doucet:2000}. This is not computationally feasible for our problem since we are learning the evolution of a large collection of latent state vectors. Instead we use the deterministic mean-field variational method for approximate posterior inference \cite{Wainwright:2008}. As we will show, this not only allows for tractable posterior calculations for the state vectors $u$, $w$ and hidden data $y$, but also allows for a transition from prior to posterior similar to the Kalman filter that retains tractability for all downstream calculations.

We approximate the posterior of Equation (\ref{eqn.jointposterior}) with the following factorized distribution,
\begin{equation}\label{eqn.Qdist}
q(u_i\<t\>,w_j\<t\>,y_{ij}\<t\>) = q(u_i\<t\>) q(w_j\<t\>) q(y_{ij}\<t\>).
\end{equation}
We derive the specific distributions in Section \ref{sec.qdistder}. This $q$ distribution approximates the posterior of each variable as being independent of all other variables, which has a significant advantage in this dynamic setting. Having defined $q$, we seek to minimizes the KL divergence between $q$ and the true posterior at time $t$ by equivalently maximizing the variational objective function \cite{Wainwright:2008},
\begin{equation}\label{eqn.objective}
\mathcal{L}_t = \mathbb{E}_q[\ln p(z_{ij}\<t\>,u_i\<t\>,w_j\<t\>,y_{ij}\<t\>)] - \mathbb{E}_q[\ln q].
\end{equation}
Since this is a non-convex function, a local maximum of $\mathcal{L}_t$ is found by optimizing the parameters of each $q$ as discussed in Section \ref{sec.coordasc}.

\subsection{Variational $q$ distribution for the prior model}\label{sec.qdistder}
As indicated in Equation (\ref{eqn.objective}), constructing the variational objective function for the event at time $t$ involves taking the expectation of the log joint likelihood and adding the entropies of each $q$. This requires defining the distribution family for $q(y_{ij}),$ $q(u_i)$ and $q(w_j)$. The optimal form for each $q$ distribution can be found by exponentiating the expected log joint likelihood holding out the $q$ distribution of interest \cite{Wainwright:2008}: $q_i \propto \exp\{\mathbb{E}_{q_{-i}}[\ln p(\cdot)]\}$. 

Temporarily ignoring the time notation, at time $t$ we can find $q(y_{ij}\<t\>)$ by calculating
\begin{equation}
 q(y_{ij}) \propto \exp\{\ln p(z_{ij}|y_{ij}) + \mathbb{E}_q[\ln p(y_{ij}|u_i,w_j)]\}.
\end{equation}
Since $p(z_{ij}\<t\>|y_{ij}\<t\>) = \mathbb{I}(y_{ij}\<t\> \in \mathcal{I}_{z_{ij}\<t\>})$, it follows that $q(y_{ij}\<t\>)$ is defined on the interval $\mathcal{I}_{z_{ij}\<t\>}$, which is the cell corresponding to class $z_{ij}\<t\>$. After taking the expectation, the optimal $q$ distribution is found to be
\begin{equation}\label{eqn.q_y}
 q(y_{ij}\<t\>) = \mathcal{T}N_{\mathcal{I}_{z_{ij}\<t\>}}\(y_{ij}\<t\>|\langle \mathbb{E}_qu_i\<t\>, \mathbb{E}_qw_j\<t\>\rangle,\sigma^2\).
\end{equation}
This is a truncated normal distribution on the support $\mathcal{I}_{z_{ij}\<t\>}$ with mean parameter $\langle\mathbb{E}_qu_i\<t\>, \mathbb{E}_qw_j\<t\>\rangle$ (defined later) and variance $\sigma^2$. 

We can follow the same procedure to find $q(u_i\<t\>)$ for time $t$. Again ignoring the time component we have
\begin{equation}
 q(u_i) \propto \exp\{\mathbb{E}_q[\ln p(y_{ij}|u_i,w_j)] + \ln p(u_i)\}.
\end{equation}
The prior on $u_i\<t\>$ is discussed below. Solving for this distribution shows that $q(u_i\<t\>)$ is a multivariate Gaussian,
\begin{equation}\label{eqn.q_x}
 q(u_i\<t\>)  =  N\(u_i\<t\>|\mu_{u_i}'\<t\>,\rSigma_{u_i}'\<t\>\),
\end{equation}
with mean and covariance parameter given below. Symmetry results in the same $q$ distribution family for $w_j\<t\>$.

\subsection{A coordinate ascent algorithm for $q(u)$, $q(w)$ and $q(y)$}\label{sec.coordasc}
We next present a coordinate ascent algorithm for updating the three $q$ distributions that appear at time $t$.\vspace{5pt}

\emph{Coordinate update of $q(y)$}:  The analytical update for the mean parameter $m_{ij}\<t\>$ of $q(y_{ij}\<t\>)$ is
\begin{equation}\label{eqn.m}
m_{ij}\<t\> = \langle\mathbb{E}_qu_i\<t\>, \mathbb{E}_qw_j\<t\>\rangle.\\
\end{equation}
The values of $\E_q u_i\<t\>$ and $\E_q w_j\<t\>$ are given below. The expectation of $y_{ij}\<t\>$ depends on the interval in which it falls according to the truncated normal. Let $l_{z_{ij}\<t\>}$ and $r_{z_{ij}\<t\>}$ be the left and right boundaries of this interval. Then defining
$$\alpha_{ij}\<t\> = \frac{l_{z_{ij}\<t\>}-m_{ij}\<t\>}{\sigma},\quad \beta_{ij}\<t\> = \frac{r_{z_{ij}\<t\>}-m_{ij}\<t\>}{\sigma},$$
we have that
\begin{equation}\label{eqn.Eprobit}
\mathbb{E}_q y_{ij}\<t\> = m_{ij}\<t\> + \sigma\frac{\phi(\alpha_{ij}\<t\>) - \phi(\beta_{ij}\<t\>)}{\Phi(\beta_{ij}\<t\>) - \Phi(\alpha_{ij}\<t\>)},
\end{equation}
where $\phi(\cdot)$ is the pdf and $\Phi(\cdot)$ the cdf of a standard normal.\vspace{5pt}

\emph{Coordinate update of $q(u)$}:  The updates for mean $\mu_{u_i}'\<t\>$ and covariance $\rSigma_{u_i}'\<t\>$ of $q(u_i\<t\>)$ are
\begin{eqnarray}
\rSigma'_{u_i} & \hspace{-5pt}=\hspace{-5pt} & \left(\rSigma_{u_i}^{-1} + (\mu_{w_j}'\mu_{w_j}'^T + \rSigma_{w_j}')/\sigma^2\right)^{-1},\nn \\\label{eqn.muUp}
\mu_{u_i}' & \hspace{-5pt}=\hspace{-5pt} & \rSigma_{u_i}'\(\E_q[y_{ij}]\mu_{w_j}'/\sigma^2 + \rSigma_{u_i}^{-1}\mu_{u_i}\).
\end{eqnarray}
All parameters above are functions evaluated at $t$. We recall that the prior mean $\mu_{u_i}\<t\> = \mu'_{u_i}[t-\Delta_{u_i}^{\<t\>}]$ and prior covariance $\Sigma_{u_i}\<t\> \approx \Sigma'_{u_i}[t-\Delta_{u_i}^{\<t\>}] + e^{a_{u_i}\<t\>}\Delta_{u_i}^{\<t\>} I$, which follows from the approximation discussed above.\vspace{5pt}

\emph{Coordinate update of $q(w)$}: Because of symmetry, the updates for $\mu_{w_j}'\<t\>$ and $\rSigma_{w_j}'\<t\>$ of $q(w_j\<t\>)$ are similar to those for $u_i$, but with the indices $u_i$ and $w_j$ switched on all means and covariances. We therefore omit these updates.

\subsection{Inferring the geometric Brownian motion $\exp\{a\<t\>\}$}
We next derive an algorithm for inferring the drift process $a\<t\>$ used in the geometric Brownian motion. We derive a point estimate of this value assuming individual $a$ for each $u_i$ and $w_j$. To update, e.g., $a_{u_i}\<t\>$ we approximate the relevant terms in $\mathcal{L}$ using a second order Taylor expansion about the point $a_{u_i}\<t-\Delta_{a_{u_i}}^{\<t\>}\>$, which is the last inferred value for this process. We recall that the form of this approximation for a general function $f(x)$ is 
$$\textstyle f(x) \approx f(x_0) + (x-x_0)f'(x_0) + \frac{1}{2}(x-x_0)^2f''(x_0),$$
with the solution at the point $x = x_0 - f''(x_0)^{-1}f'(x_0)$. In our case, $f = \mathbb{E}_q[\ln p(u_i\<t\>,a_{u_i}\<t\>)]$. 

We can efficiently update $a_{u_i}\<t\>$ using this Taylor expansion as follows: Let $\Sigma'_{u_i}[t-\Delta_{u_i}^{\<t\>}] = Q\Lambda Q^T$ be the eigendecomposition of the posterior covariance of $u_i$ at the previous observation. (This needs to be done only once for the updates at this time point.) Also, let $$v = (\mu'_{u_i}\<t\> - \mu_{u_i}\<t\>)^TQ,\quad M = Q^T\Sigma'_{u_i}\<t\> Q.$$ Since this depends on the updated posterior mean and covariance of $u_i$, $v$ and $M$ should be recomputed after each update of $q(u_i)$. Let $\lambda_d$ be the $d$th eigenvalue of $\Sigma'_{u_i}[t-\Delta_{u_i}^{\<t\>}]$ and define
$$\eta_d = \frac{e^{a_{u_i}\<t\>}\Delta_{a_{u_i}}}{\lambda_d + e^{a_{u_i}\<t\>}\Delta_{a_{u_i}}}.$$
Then using a second order Taylor expansion of the objective function at the previous point $a_{u_i}\<t-\Delta_{a_{u_i}}\>$ gives the following first and second derivatives with respect to $a_{u_i}$,
\begin{eqnarray}
f' &\hspace{-2pt}=\hspace{-2pt}& \textstyle -\frac{a\<t\> - a\<t-\Delta_a\>}{c\Delta_a} - \frac{1}{2}\sum_d \eta_d(1-\frac{v_d^2 + M_{d,d}}{\lambda_d + e^{a\<t\>}\Delta_a}),\nn\\
f'' &\hspace{-2pt}=\hspace{-2pt}&\textstyle -\frac{1}{c\Delta_a} - \frac{1}{2}\textstyle\sum_d \eta_d \left(1-\eta_d\right)\nn\\
 &&+ \textstyle\frac{1}{2}\textstyle\sum_d \eta_d \left(1 - 2\eta_d\right)\frac{v_d^2 + M_{d,d}}{\lambda_d + e^{a\<t\>}\Delta_a}.\label{eqn.a}
\end{eqnarray}
We have removed the subscript dependency on $u_i$ above. We then use the closed form equation from the Taylor expansion to update $a_{u_i}\<t\>$. The update for $a_{w_j}\<t\>$ has the same form, but with the relevant variational parameters and eigendecomposition replacing those above.

\subsection{Simplifications to the model and predictions of new data}\label{sec.ckf_y} 
When $y_{ij}\<t\>$ is observed only a slight modification needs to be made to the algorithm above. In this case $q(y)$ is no longer necessary and in the update for each $q(u)$ and $q(w)$ the term $\mathbb{E}_qy_{ij}\<t\>$ can be replaced with the observed value.

Predictions using the CKF require an approximation of an intractable integral. One approximation is to use Monte Carlo methods. Suppressing the time $t$, the expectation we are interested in taking is 
$\upsilon_{ij} := \E_q[\Phi(\langle u_{i}, w_{j}\rangle/\sigma)]$  We can approximate $\upsilon_{ij}$ using $S$ samples from these distributions,
\begin{equation}
\hat{\upsilon}^{(S)}_{ij} = \frac{1}{S}\sum_{s=1}^S \Phi(\langle u_{i}^{(s)}, w_{j}^{(s)}\rangle/\sigma),
\end{equation}
where $u_{i}^{(s)} \stackrel{iid}{\sim} q(u_{i}\<t\>)$ and $w_{j}^{(s)} \stackrel{iid}{\sim} q(w_{j}\<t\>)$. This gives an unbiased estimate of the expectation. A faster approximation is to directly use the means from $q$ for $u_i\<t\>$ and $w_j\<t\>$.

\section{Experiments}\label{sec.experiments}
We evaluate the performance of the collaborative Kalman filter on the following three data sets:
\begin{itemize}
 \item The Netflix data set containing 100 million movie ratings from 1999 to 2006. The movies are rated from 1 to 5 and ratings are distributed across roughly 17K movies and 480K users. 
 \item The MovieLens data set containing 10 million movie ratings from 1995 to 2009. The ratings are given on a half star scale from 0.5 to 5 and are distributed across 10K movies and 71K users.
  \item Stock returns data measured at opening and closing times for 433 companies from the AMEX exchange, 2,774 companies from NASDAQ and 3,273 companies from the NYSE for a total of 6,480 stocks and 39.1 million total measurements from 1962--2014.\footnote{Data downloaded from \url{http://ichart.finance.yahoo.com/}}
\end{itemize}
The model requires setting some important parameters: The dimension $d$ of $u_i$ and $w_j$, the standard deviation of the probit model $\sigma$, the partition widths $r_k-l_k$, and the drift parameter $c$ for the geometric Brownian motion $a\<t\>$. We discuss these below for the respective problems.

 \begin{figure}[t!]
\subfigure{\includegraphics[width=.235\textwidth]{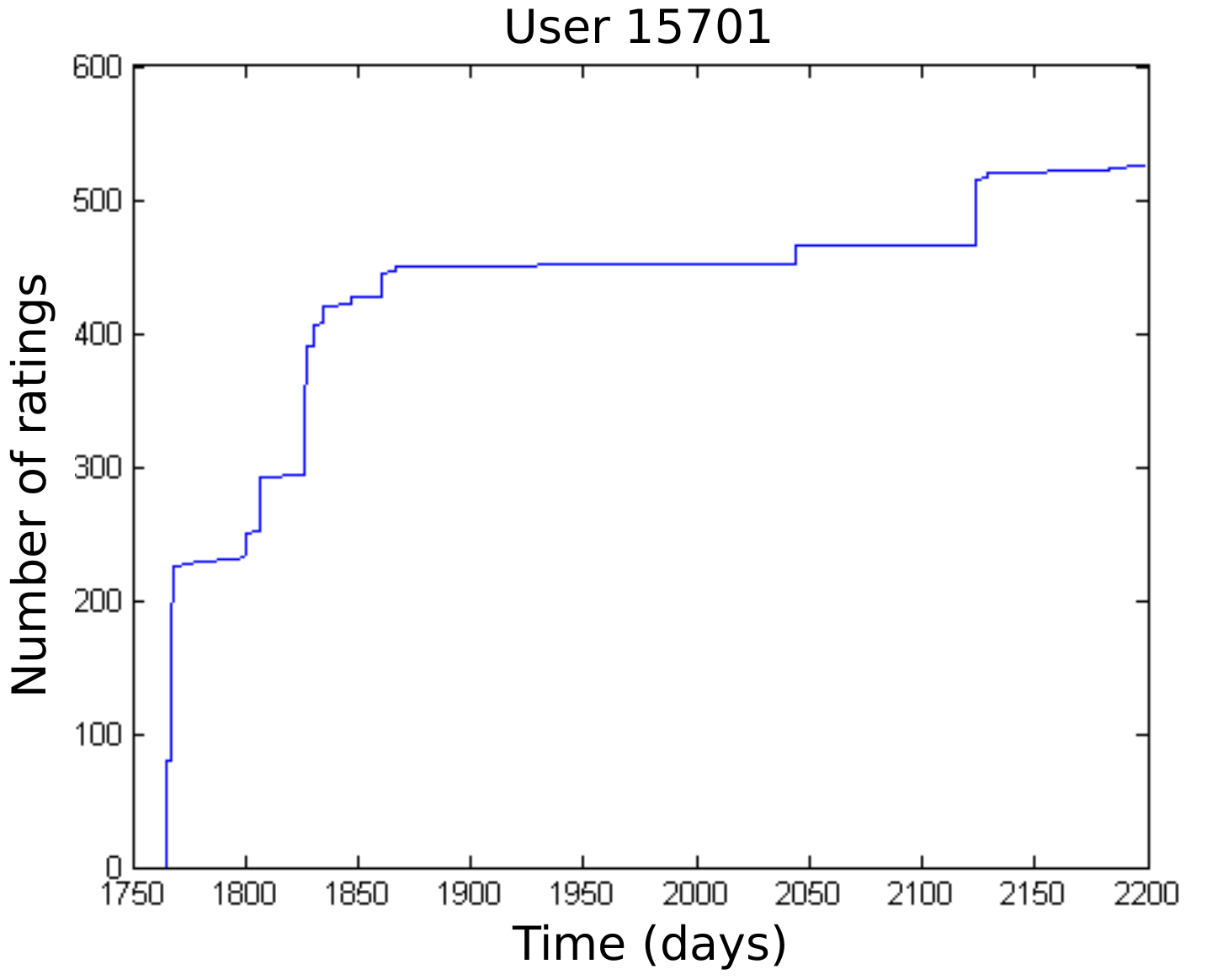}}
\subfigure{\includegraphics[width=.235\textwidth]{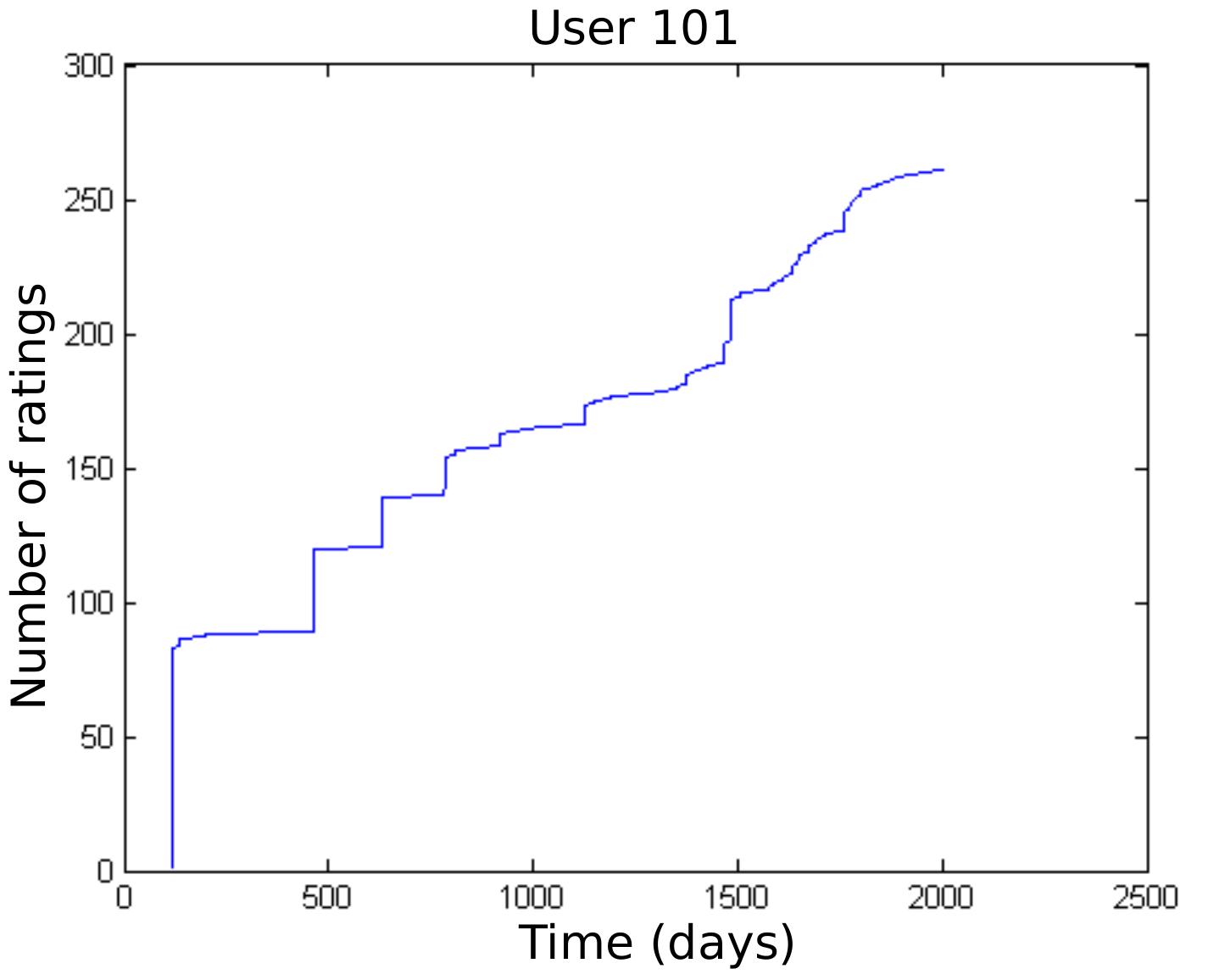}}
\caption{Cumulative total of movies rated by two users from the Netflix data set. (left) This user rates large batches of movies in a few sittings. Though the dynamics won't be captured by the model, we still can perform sequential inference for this user to make predictions. (right) A more incremental rating pattern that has dynamic value. }\label{fig.twousers}
\end{figure}

\subsection{Movie rating prediction}
We first present results on dynamic modeling of the Netflix and MovieLens data sets. For these problems, we tried several different dimension settings ($d=10,20,30$). We learned a converging $a_u$ shared among users and $a_w$ shared by movies by setting $c=0$ since we do not assume there to be fundamental shifts in the overall user or movie landscape. We set $\sigma$ to be the value that minimizes the KL divergence between the probit and logistic link functions, which we found to be approximately $\sigma = 1.76$. We randomly initialized the mean of the priors on each $u_i$ and $w_j$ at time zero, and set the covariance equal to the identity matrix. For the partition width, we set $r_k-l_k = \sigma$ for Netflix and $r_k-l_k = \sigma/2$ for MovieLens, which accounts for the half vs.\ whole star rating system. We compare with several algorithms:
\begin{enumerate}
 \item Online VB-EM: The non-drift special case of the CKF with $\exp\{a\<t\>\} = 0$.
 \item Batch VB-EM: The batch variational inference version of (1) that uses all ratings when updating $u_i$ and $w_j$.
 \item Batch MAP-EM: A version of probabilistic matrix factorization (PMF) \cite{Salakhutdinov:2007} that uses the probit model as above. Point estimates of $u_i$ and $w_j$ are learned and the hidden data $y_{ij}$ is inferred using EM.
 \item BPMF: Bayesian PMF \cite{Salakhutdinov:2008} without a probit model.
 \item M$^3$F: Mixed-membership matrix factorization \cite{Mackey:2010}.
\end{enumerate}
The zero-drift version of this model is an online sequential method for Bayesian inference that is similar to other ``big data'' extensions of the variational inference approach \cite{Broderick:2013,Hoffman:2013}. The difference between this and the batch model is that we only process each rating once with the online algorithm, while with batch inference we iterate over users and movies several times processing all data in a single iteration, as is more typically done.

\begin{figure}[t!]
\subfigure[Netflix]{\includegraphics[width=.235\textwidth]{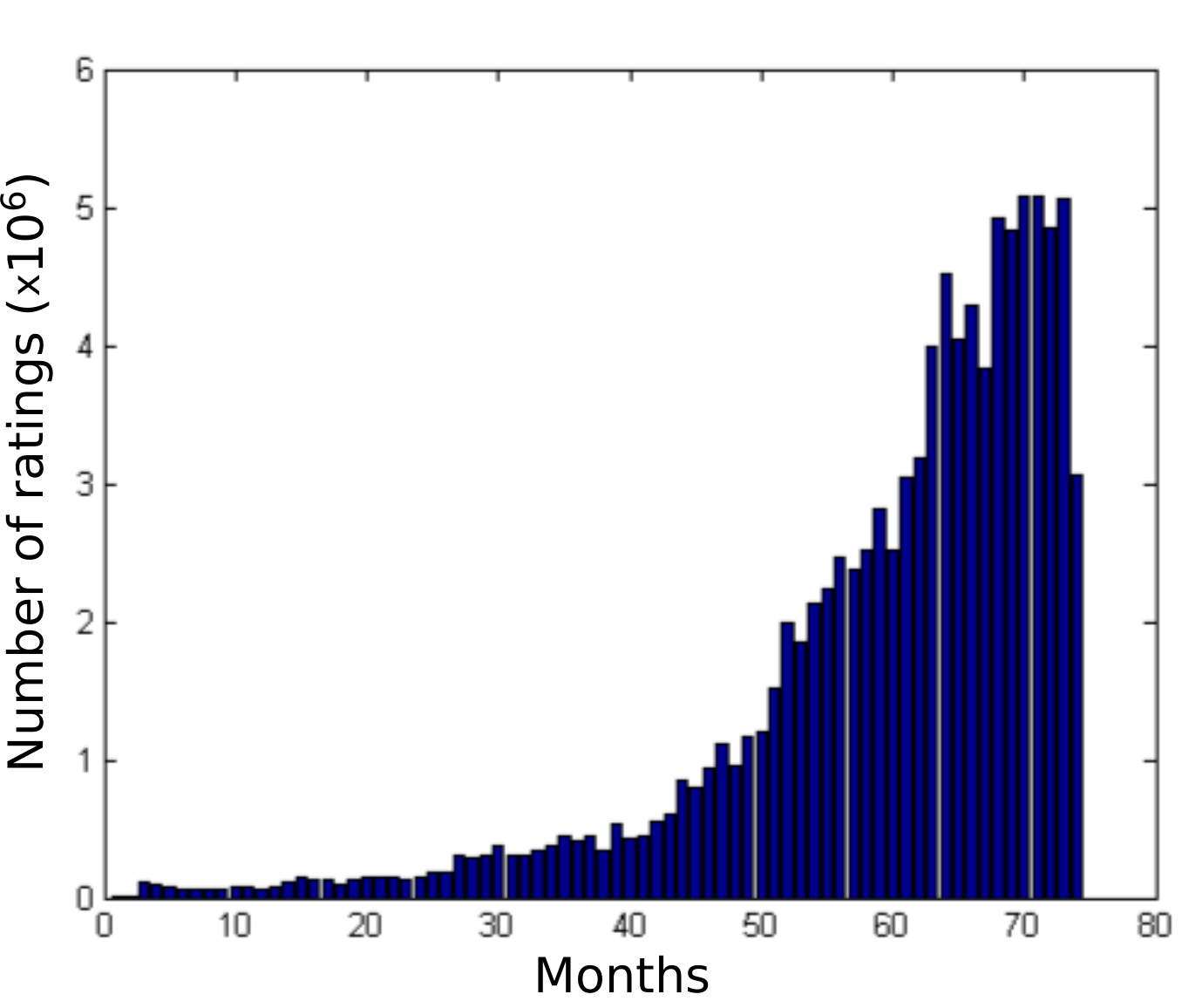}}
\subfigure[MovieLens]{\includegraphics[width=.24\textwidth]{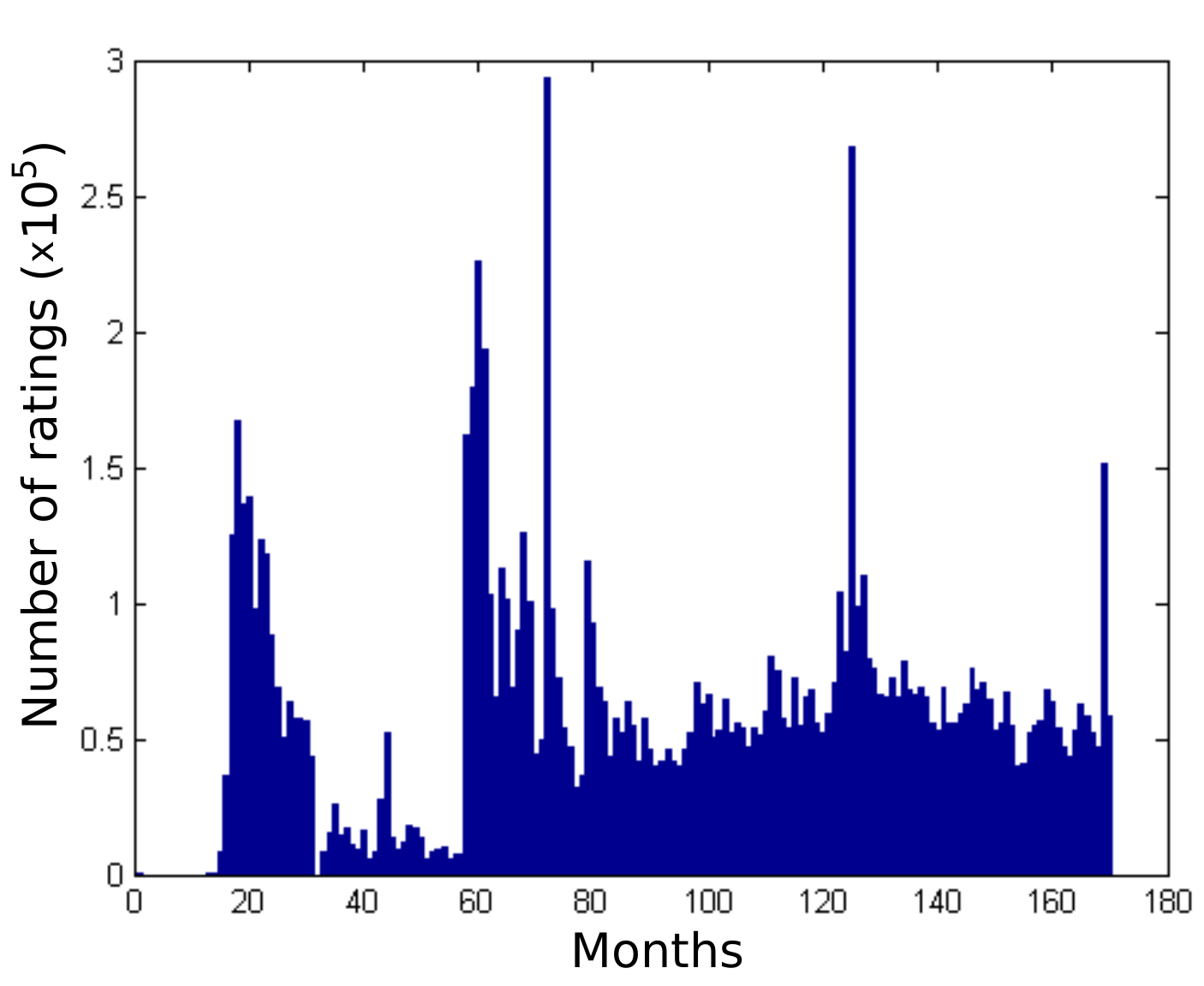}}
\caption{Histograms of the total number of ratings within a month over the course of each data set.}\label{fig.moviehist}
\end{figure}

The ability of our model to exploit dynamic information depends on how dynamic the data is. For example, in Figure \ref{fig.twousers} we show the cumulative number of ratings for two users as a function of time. With the first user, we don't expect to have a dynamic benefit, but we do for the second user. However, as an online algorithm we note that the CKF still can make useful predictions in both cases; in the limiting case that all users rate all movies in a given instant, the CKF will simply reduce to the online zero-drift model. In Figure \ref{fig.moviehist} we show the monthly ratings histogram for both data sets, which gives a sense of the dynamic information from the movie perspective.

\begin{table}\normalsize
\caption{RMSE results for the Netflix 100 million and MovieLens 10 million data sets. Comparisons show an advantage to modeling the dynamic information within the data.}\label{tab.rmse}\centering
\begin{threeparttable}
\begin{tabular}{ llcc }
\toprule
Model & ~Size~~~~~ & ~~Netflix~~ & ~MovieLens~  \\
\midrule
CKF & $d = 10$ & 0.8540  & 0.7726\\
    & $d = 20$ & 0.8534 & 0.7654\\
    & $d = 30$ & 0.8540 & 0.7635\vspace{2pt}\\
Online VB-EM & $d = 10$ & 0.8682 & 0.7855\\
    & $d = 20$ & 0.8707 & 0.7805\\
    & $d = 30$ & 0.8668 & 0.7786\vspace{2pt}\\
Batch VB-EM & $d = 10$ & 0.8825 & 0.7996\\
    & $d = 20$ & 0.8688 & 0.7896\\
    & $d = 30$ & 0.8638 & 0.7865\vspace{2pt}\\
Batch MAP-EM & $d = 10$ & 0.9277 & 0.9133\\
    & $d = 20$ & 0.9182 & 0.9113\\
    & $d = 30$ & 0.9143 & 0.9133\vspace{2pt}\\
BPMF & $d = 30$ & 0.9047 & 0.8472\vspace{2pt}\\
M$^3$F & $d = 30$ & 0.9015 & 0.8447 \\
\bottomrule
\end{tabular}
\end{threeparttable}
\end{table}

We use the RMSE as a performance measure, which we show for all algorithms in Table \ref{tab.rmse}. For the batch algorithms, we randomly held out 5\% of the data for testing to calculate this value. We ran multiple trials and found that the standard deviation for these algorithms was small enough to omit. (We discuss training/testing for the online algorithms below.) We observe that our algorithm outperforms the other baseline algorithms, and so dynamic modeling of user preference does indeed given an improvement in rating prediction. This is especially evident when comparing the CKF with Online VB, the only difference between these algorithms being the introduction of a drift in the state space vectors. 

We also observe the improvement of the variational framework in general. For example, by comparing Batch VB with PMF EM, we see that variational inference provides an improvement over a MAP EM implementation, which models a point estimate of $u_i$ and $w_j$. Both methods use a latent variable $y_{ij}$ in a probit function for the observed rating, but the variational approach models the uncertainty in these state space vectors, and so we see that a fully Bayesian approach is helpful. We also found in our experiments that treating the rating $z_{ij}\<t\>$ as being generated from a probit model and learning a latent $y_{ij}$ is important as well, which we observed by comparing PMF EM with the original PMF algorithm \cite{Salakhutdinov:2007}.\footnote{We omit these PMF results in the table, which we note gave RMSE results over one as can be seen in the original paper. We also note that the PMF algorithm is the non-dynamic version of \cite{Sun:2012}.}

\begin{figure}
\subfigure[Netflix~~~~~~~]{\includegraphics[width=.235\textwidth]{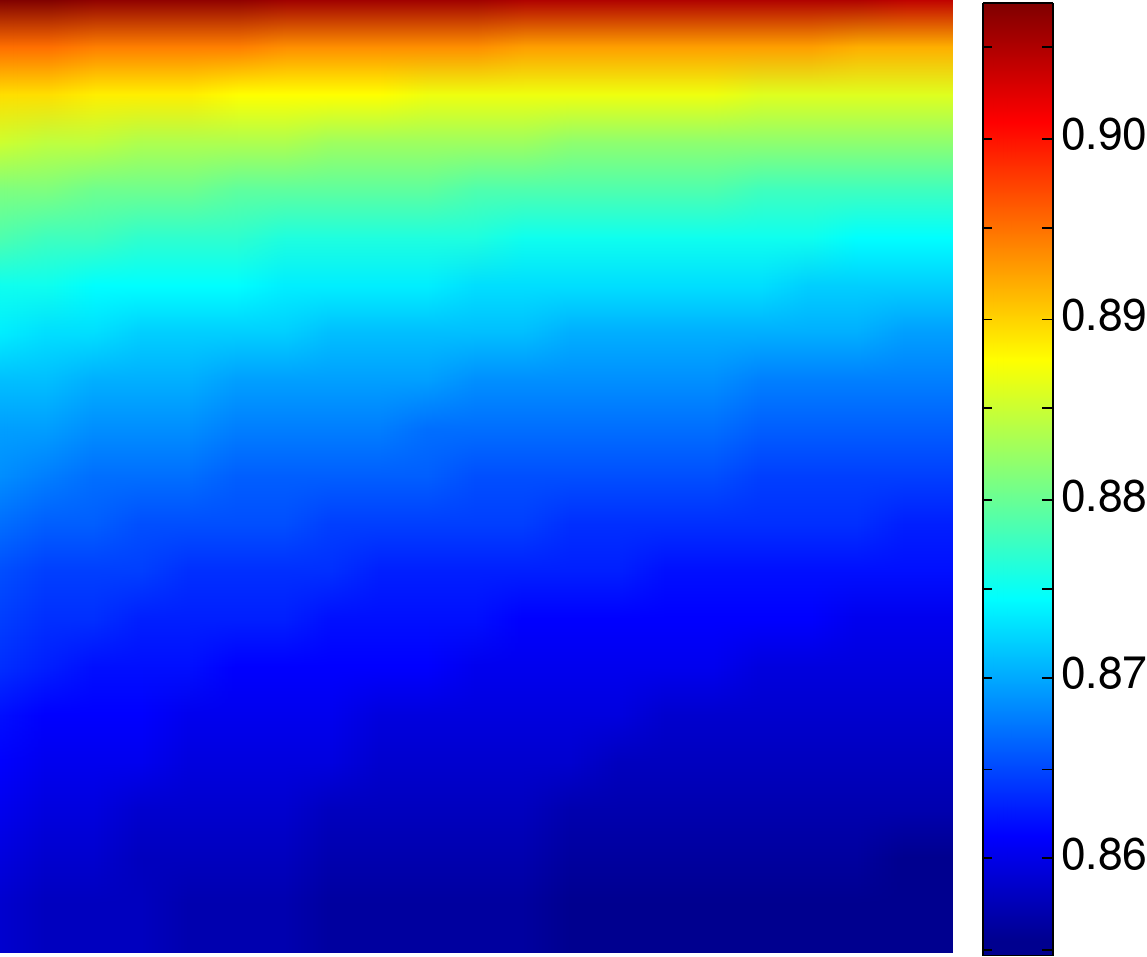}}
\subfigure[MovieLens~~~~~~~]{\includegraphics[width=.235\textwidth]{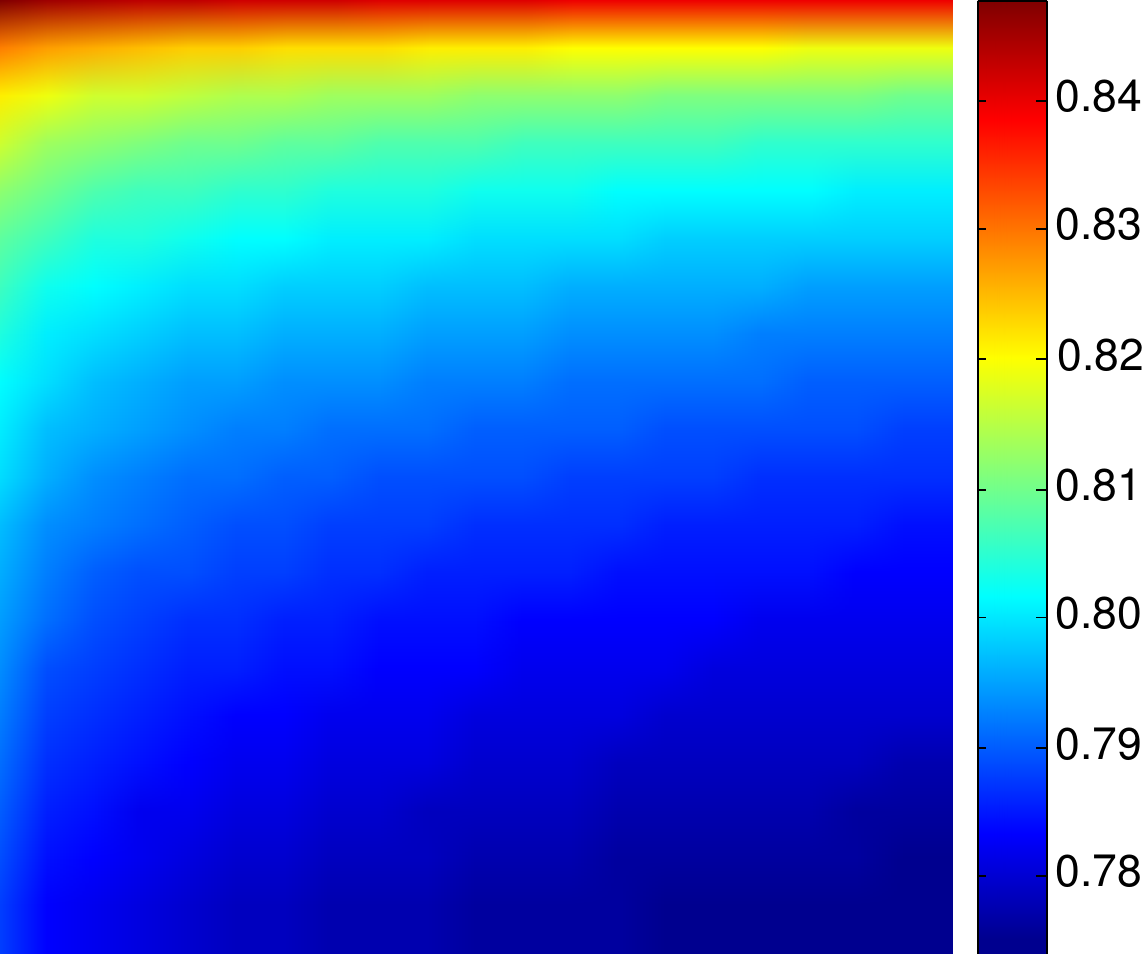}}
\caption{The RMSE as a function of number of ratings for a user and movie. The $(m,n)$ entry contains the RMSE calculated over user/movie pairs where the user has rated at least $10(m-1)$ movies and the movie has been rated at least $10(n-1)$ times. The value shown in the lower-right corner is $200+$ ratings each, which we use in Table \ref{tab.rmse}.}\label{fig.rmseheat}
\end{figure}

Calculating the RMSE requires a different approach between the online and static models. To calculate the RMSE for the two dynamic models---CKF and the online model---we do not use the test set for the batch models, but instead make predictions of every rating in the data set {\it before} using the observed rating to update the model. This gives a more realistic setting for measuring performance of the online models, especially for the CKF where we're interested in predicting the user's rating {\it at that time}. Therefore, we must choose which predictions to calculate the RMSE over, since clearly it will be bad for the first several ratings for a particular user or movie. By looking at the users and movies that appear in the testing sets of the batch models we found that for Netflix each user in the test set had an average of 613 ratings in the training set and each movie in the test set had 53,395 ratings in the training set. For MovieLens this was 447 per user and 7,157 per movie, meaning that in both 
cases a 
substantial amount of data was used to learn locations in training before making predictions in testing. Therefore, to calculate our RMSE of the online models we disregard the prediction if 
either the user or movie has under 200 previous ratings. This arguably still puts the RMSE for our model at a disadvantage, but we noticed that the value didn't change much with values larger than 200. We show the RMSE as a function of this number in Figure \ref{fig.rmseheat}.

\begin{figure}\centering
 \includegraphics[width=.45\textwidth]{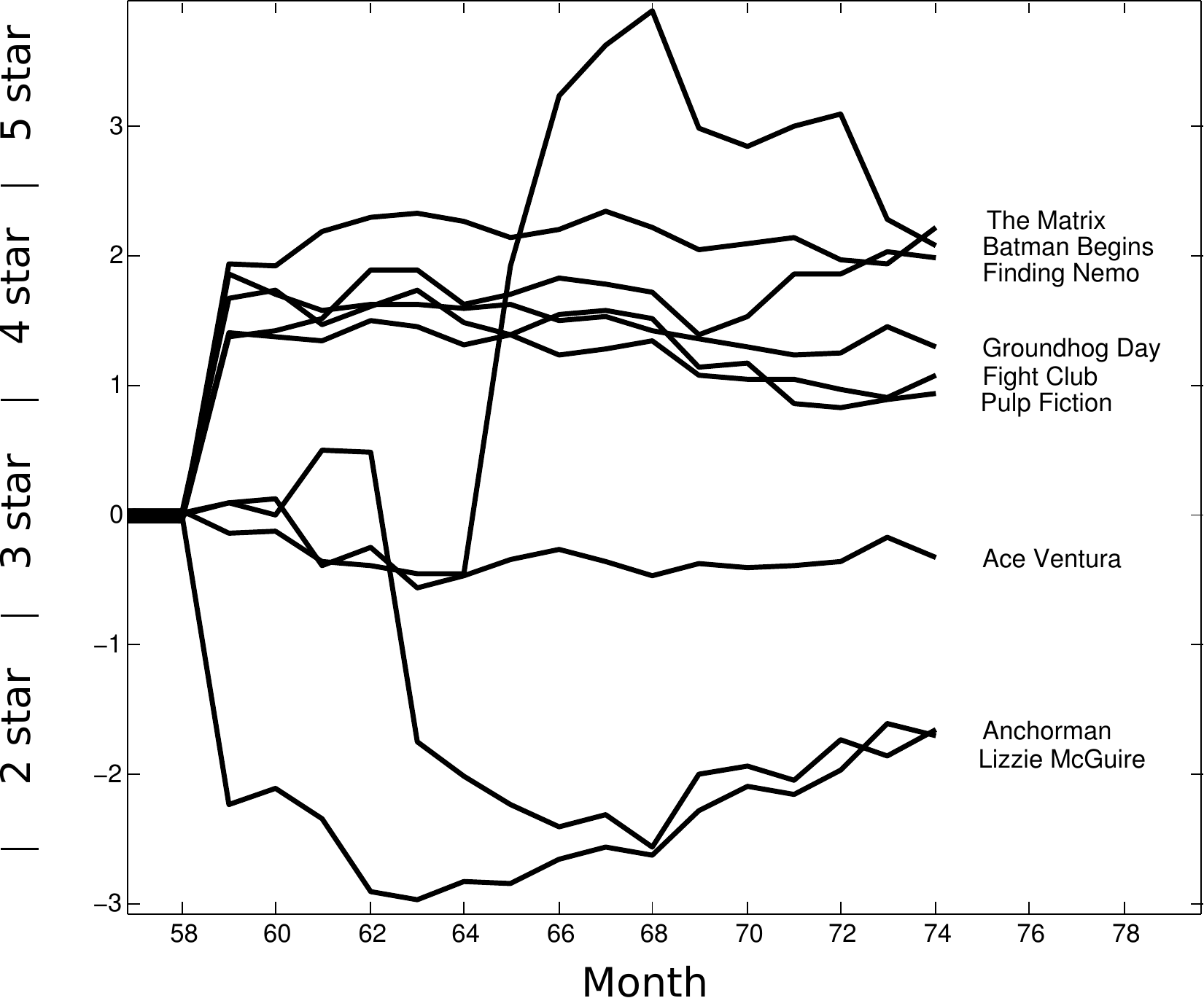}
 \caption{An example of a user drift over time as seen through predicted ratings for several movies from the Netflix data set. The y-axis is the latent variable space, which we partition according to star rating as indicated.}\label{fig.userdrift}
\end{figure}

In Figure \ref{fig.userdrift} we show the dynamics of an individual user by plotting the predicted rating for a set of movies as a function of time. We can see an evolving preference in this plot--for example a strong interest in Batman Begins that then slightly decreases with time, while interest in The Matrix begins to increase toward the end. Also, while there's some interest in Anchorman at the beginning, the user then quickly becomes uninterested. In most cases, we observed no clear evolution in movie preference for a user. We consider to be intuitively reasonable since we argue that few people fundamentally change their taste. However, being able to find those individual users or movies that do undergo a significant change can have an impact on learning the latent vectors for \emph{all} users and movies since the model is \emph{collaborative}, and so we argue that modeling time evolution can be valuable even when the actual percentage of dynamically changing behaviors is small.

\subsection{Stock Data}
\begin{figure}
 \includegraphics[width=.45\textwidth]{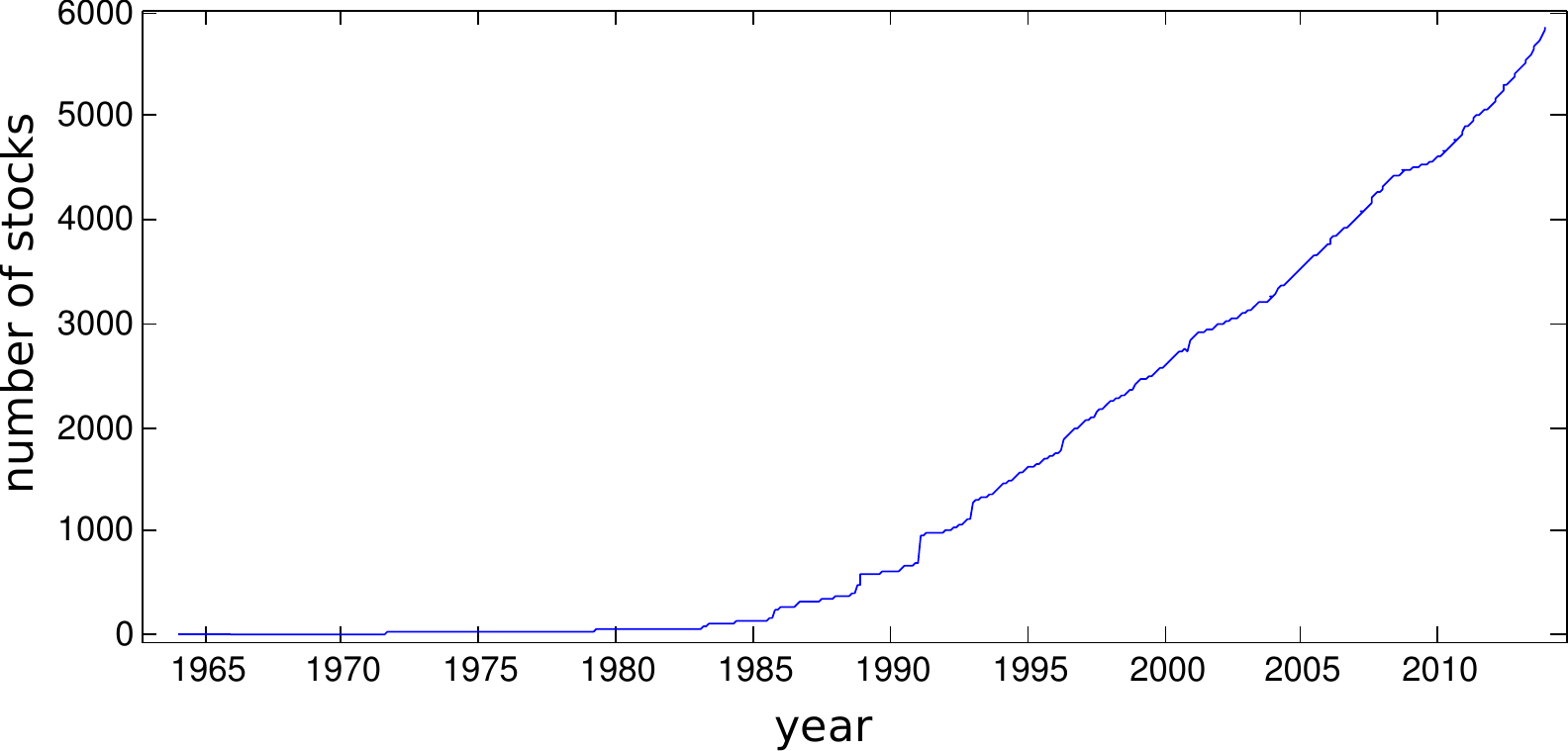}
 \caption{The number of actively traded stocks as a function of time.}\label{fig.num_stocks}
\end{figure}

We next present a qualitative evaluation of our model on a stock returns data set. For this problem, the value of $y_{ij}\<t\>$ is observed since it is treated as the continuous-valued stock price, so we can model it directly as discussed in section \ref{sec.ckf_y}. We set the latent dimension $d=5$, but observed similar results for higher values. We learned stock-specific drift Brownian motions $a_{u_i}\<t\>$ and set $c_u = 5\times10^{-2}$, which we found to be a good setting through parameter tuning since the learned $a_{u_i}\<t\>$ were not too smooth, but still stable. We also observe that, for this problem, there is only one state vector corresponding to $w$, which we refer to as a ``state-of-the-world'' (SOW) vector. Therefore, this problem falls more within the traditional Kalman filtering setting discussed in Section \ref{sec.kalmanfilter}. For the SOW vector, we set $c_w = 0$ to learn a converging value of $a_w\<t\>$, which we note was $a_w\<t\> \rightarrow -11.7$. For the noise 
standard deviation we set $\sigma = 0.01$, which enforce that the state space vectors track $y_{ij}\<t\>$ closely. (We again note that $j=1$ for this problem.) We plot the number of active stocks by year in Figure \ref{fig.num_stocks} where we see that as time goes by, the number of stocks actively being traded increases significantly.

\begin{figure}
 \subfigure{\includegraphics[width=.25\textwidth]{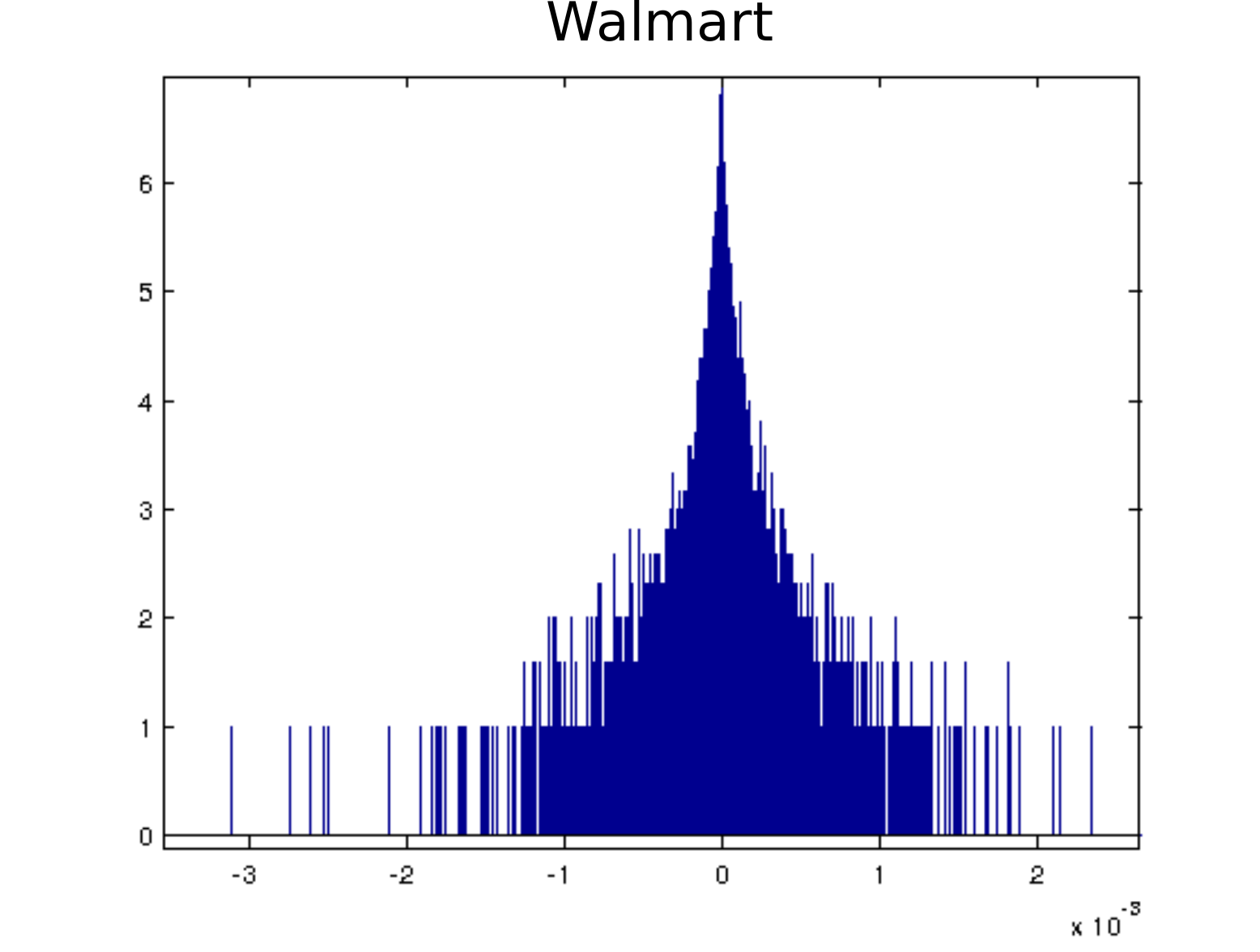}}\hspace{-100pt}
  \subfigure{\includegraphics[width=.25\textwidth]{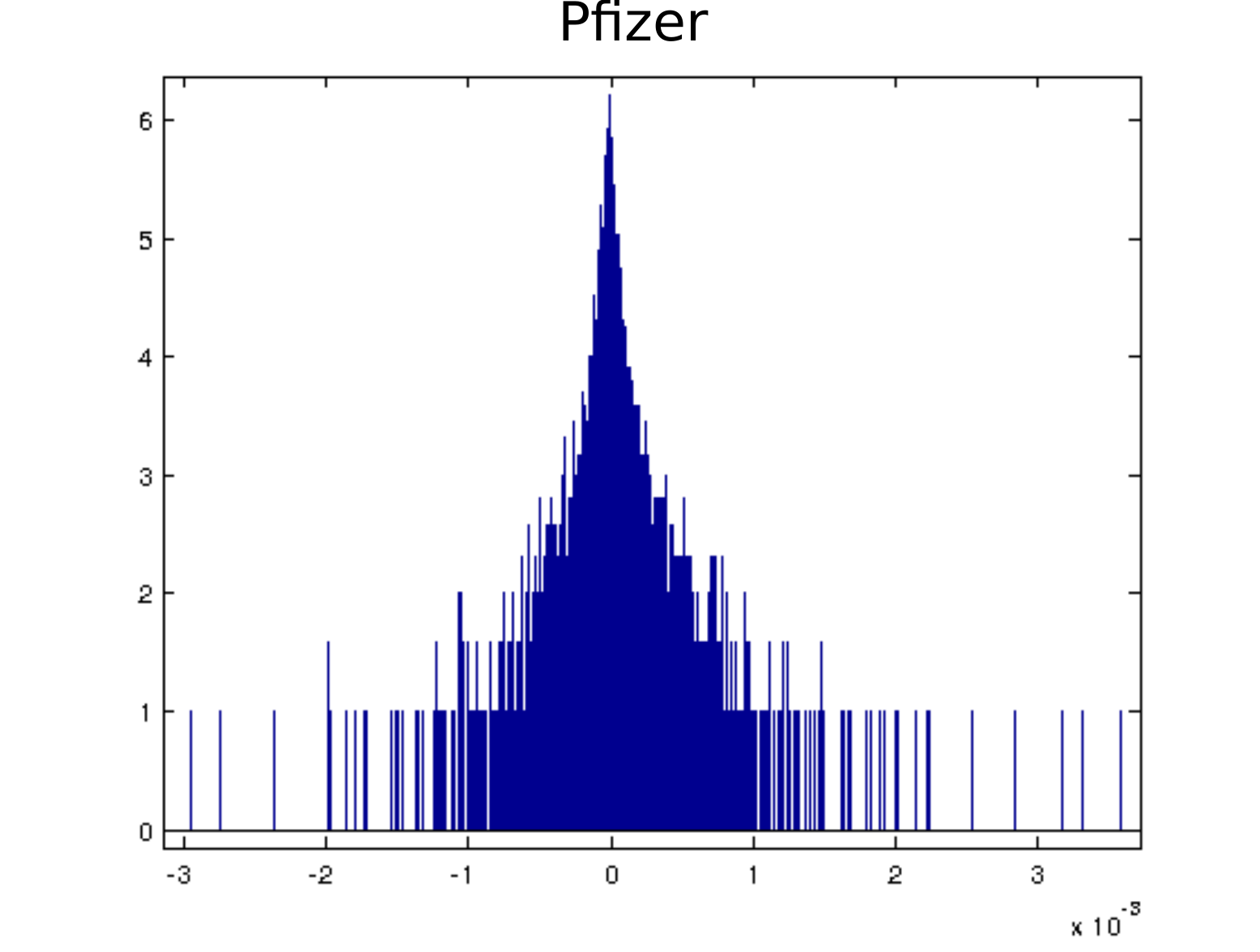}}
 \subfigure{\includegraphics[width=.25\textwidth]{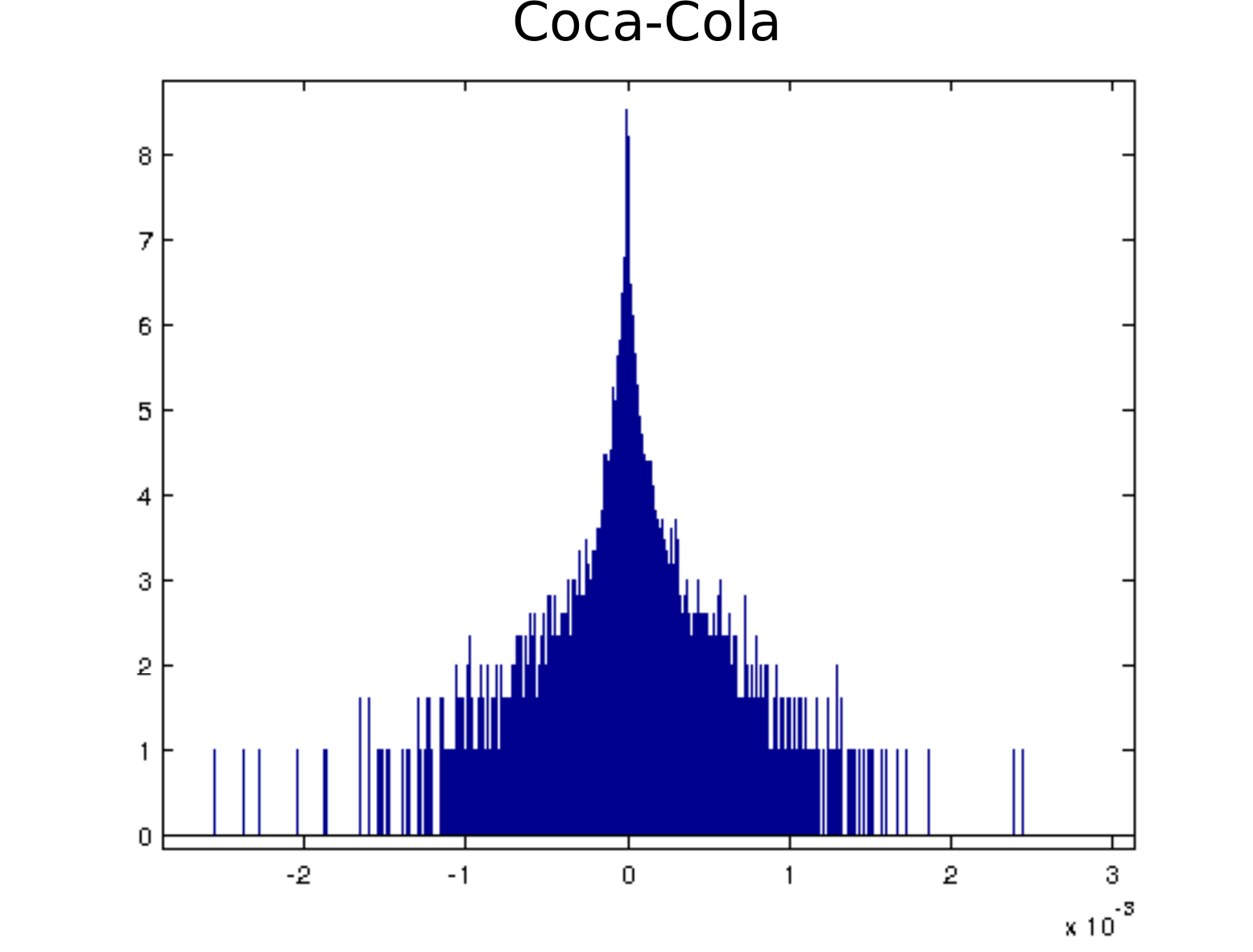}}\hspace{-15pt}
 \subfigure{\includegraphics[width=.25\textwidth]{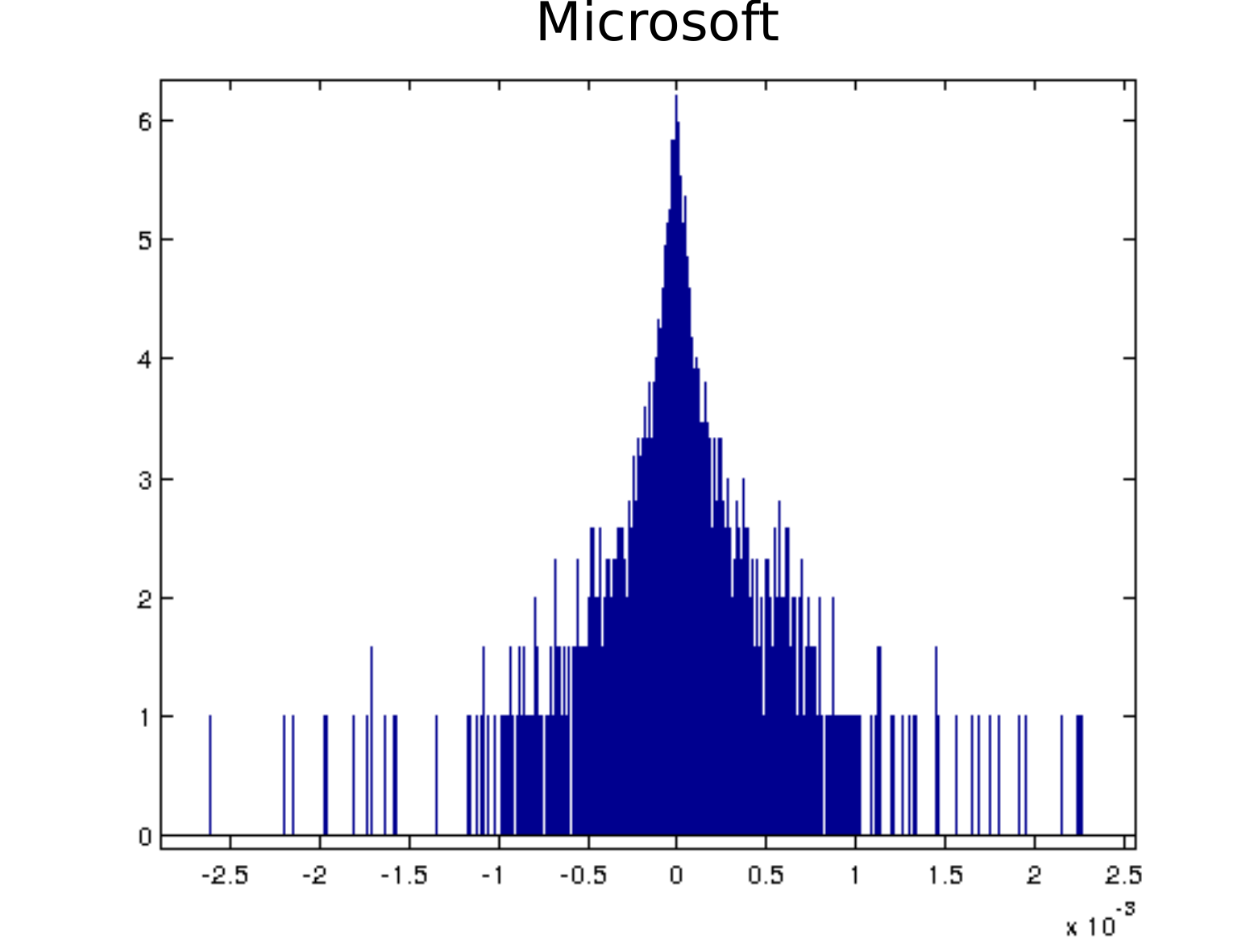}}
 \caption{A histogram of the tracking errors of the CKF for four stocks using the mean of each $q$ distribution (in log$_2$ scale). The tracking performance is very accurate using a 5 dimensional latent space (order $10^{-3}$). These result are typical of all histograms.}\label{fig.err_hist}
\end{figure}

We first assess the tracking ability of our model. The ability to accurately track the stock prices indicates that the latent structure being learned is capturing meaningful information about the data set. We show these results as error histograms on log$_2$ scale in Figure \ref{fig.err_hist} for four companies (tracking was very accurate and could not be distinguished visually from the true signal) and mention that these results are representative of all tracking performances. We see from these plots that the prediction errors of the stock prices are small, on the order of $10^{-3}$, and so we can conclude that our five dimensional state space representation for $u_i$ and $w$ is sufficient to capture all degrees of freedom in time across the 6,480 stocks.

\begin{figure}
  \subfigure[Stock price]{\includegraphics[width=.46\textwidth]{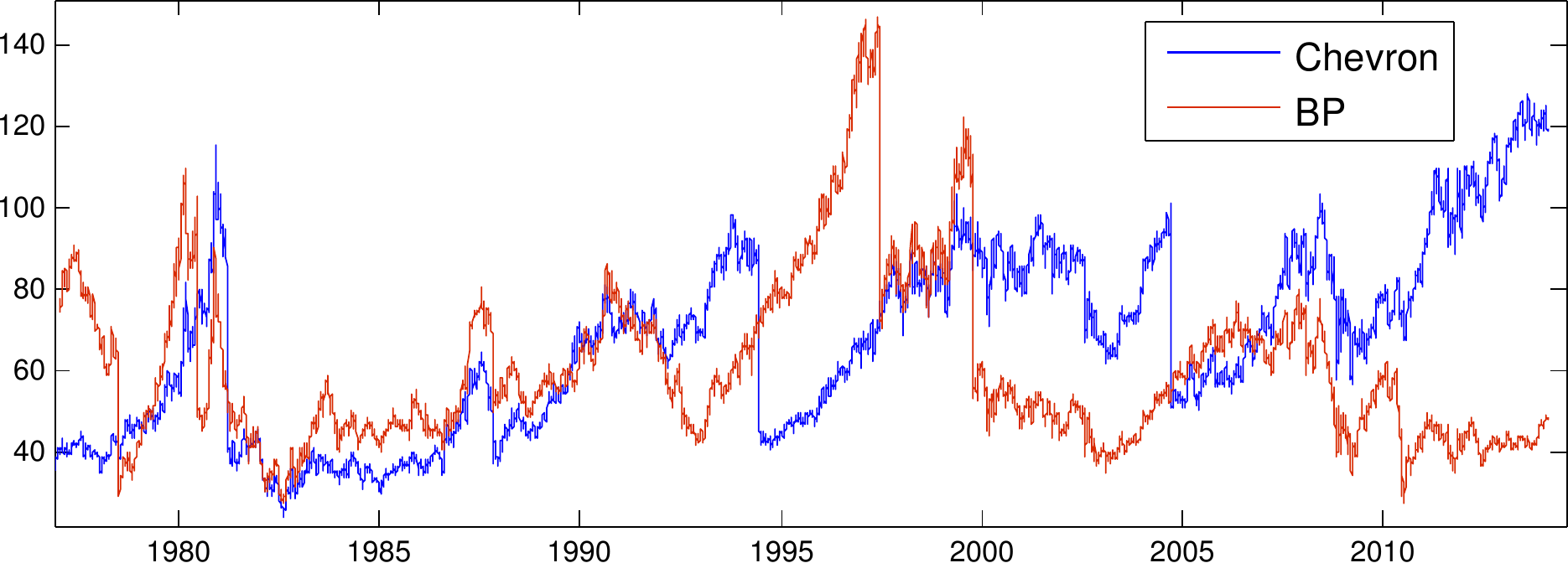}}
  \subfigure[Brownian motion volatility parameter $a_{u_i}\<t\>$.]{\includegraphics[width=.46\textwidth]{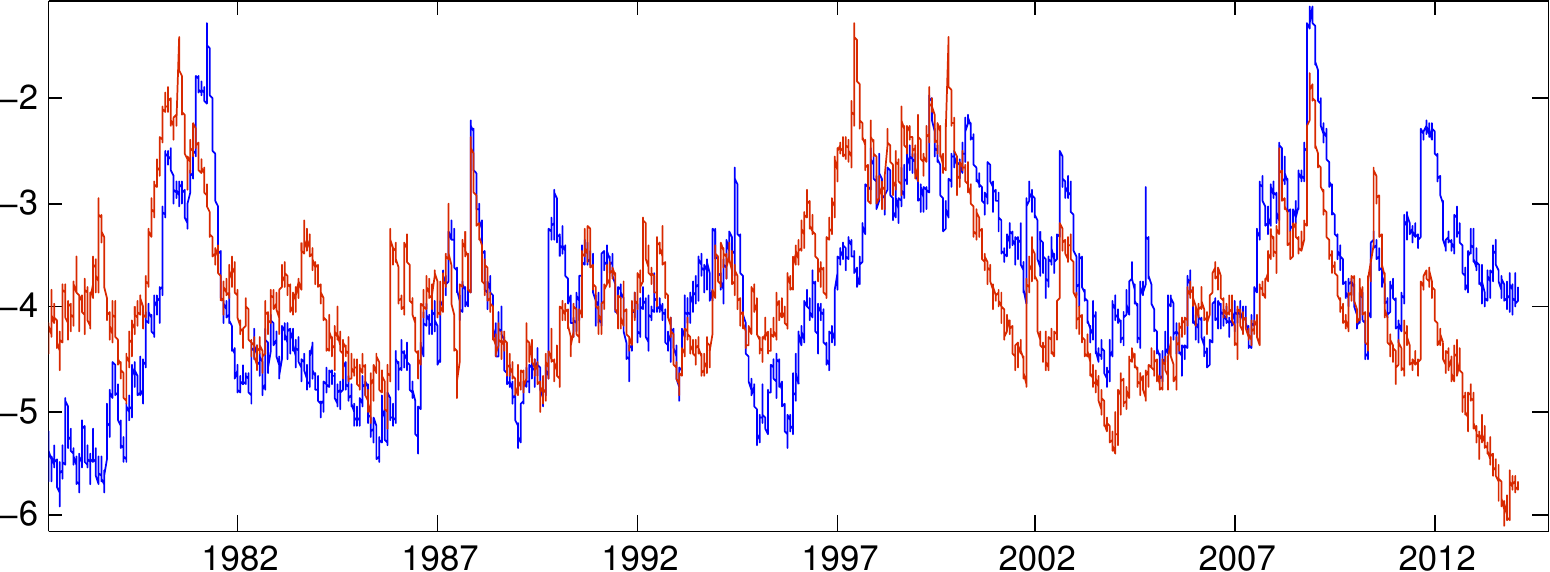}}
  \caption{(a) The historical stock price for two oil companies, BP and Chevron. (b) The log drift Brownian motion ($a_{u_i}\<t\>$) indicating the volatility of each stock. We see that though the stock prices are different, the volatility of both oil companies share the same shape since they are closely linked in the market.}\label{fig.oil}
\end{figure}

We next look more closely at the Brownian motions $a_{u_i}\<t\>$ learned by the model to capture the volatility of the stock prices. In Figure \ref{fig.oil}(a) we show the historical stock price for two oil companies, BP and Chevron. Below this in Figure \ref{fig.oil}(b) we show the respective functions $a_{u_i}$ learned for these stocks. We see that the volatility of both of these oil companies share similar shapes since they are closely linked in the market. For example in the early 80's, late 90's and late 2000's, oil prices were particularly volatile. This is modeled by the increase of the geometric Brownian motion $\exp\{a_{u_i}\<t\>\}$, which captures the fact that the state vectors are moving around significantly in the latent space during this time to rapidly adjust to stock prices.

\begin{figure*}
 \subfigure{\includegraphics[width=.195\textwidth]{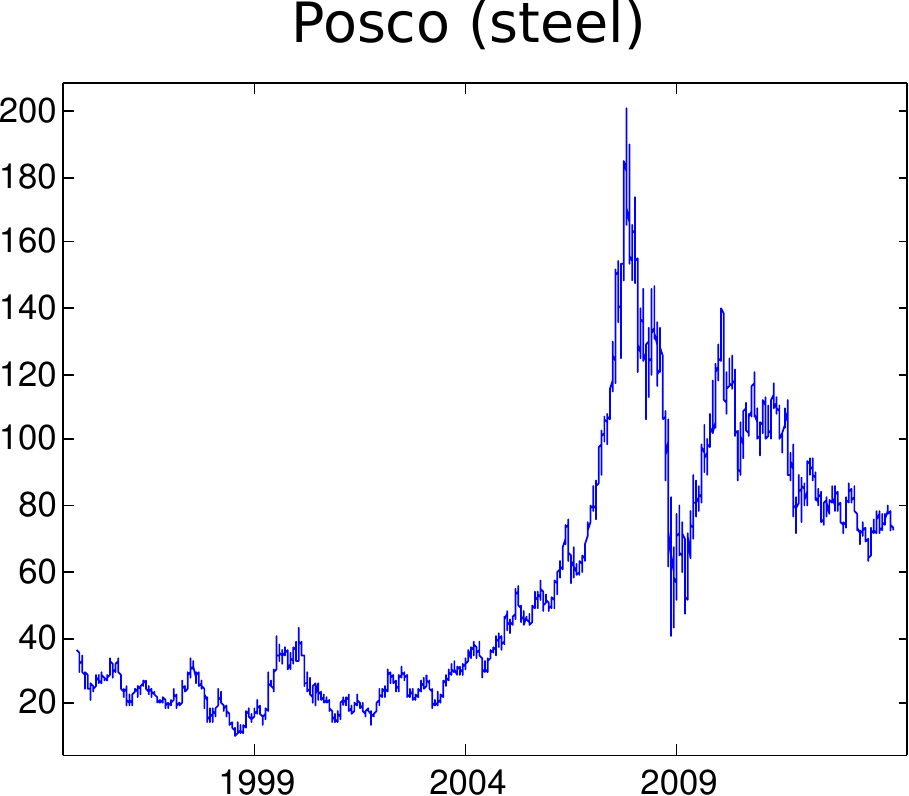}}
  \subfigure{\includegraphics[width=.195\textwidth]{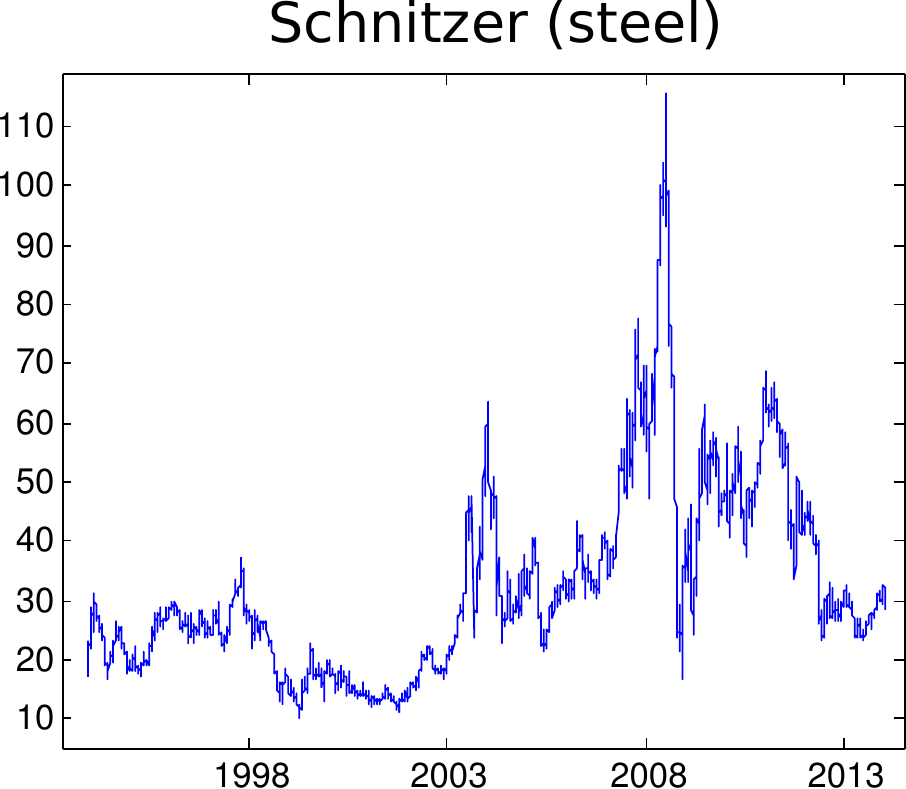}}
 \subfigure{\includegraphics[width=.195\textwidth]{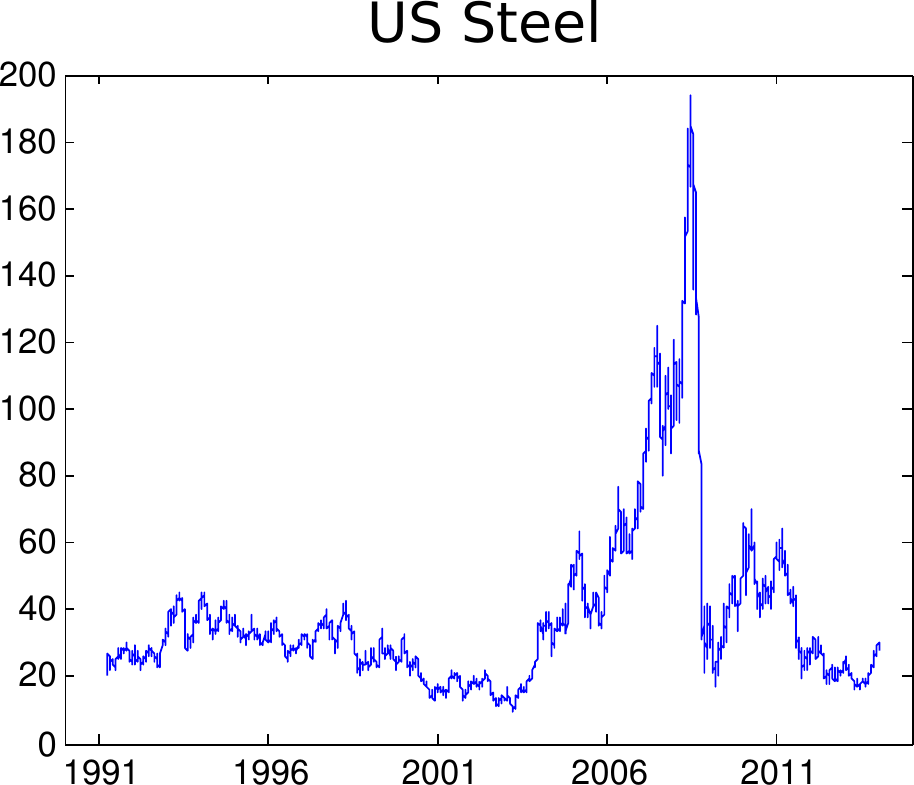}}
  \subfigure{\includegraphics[width=.195\textwidth]{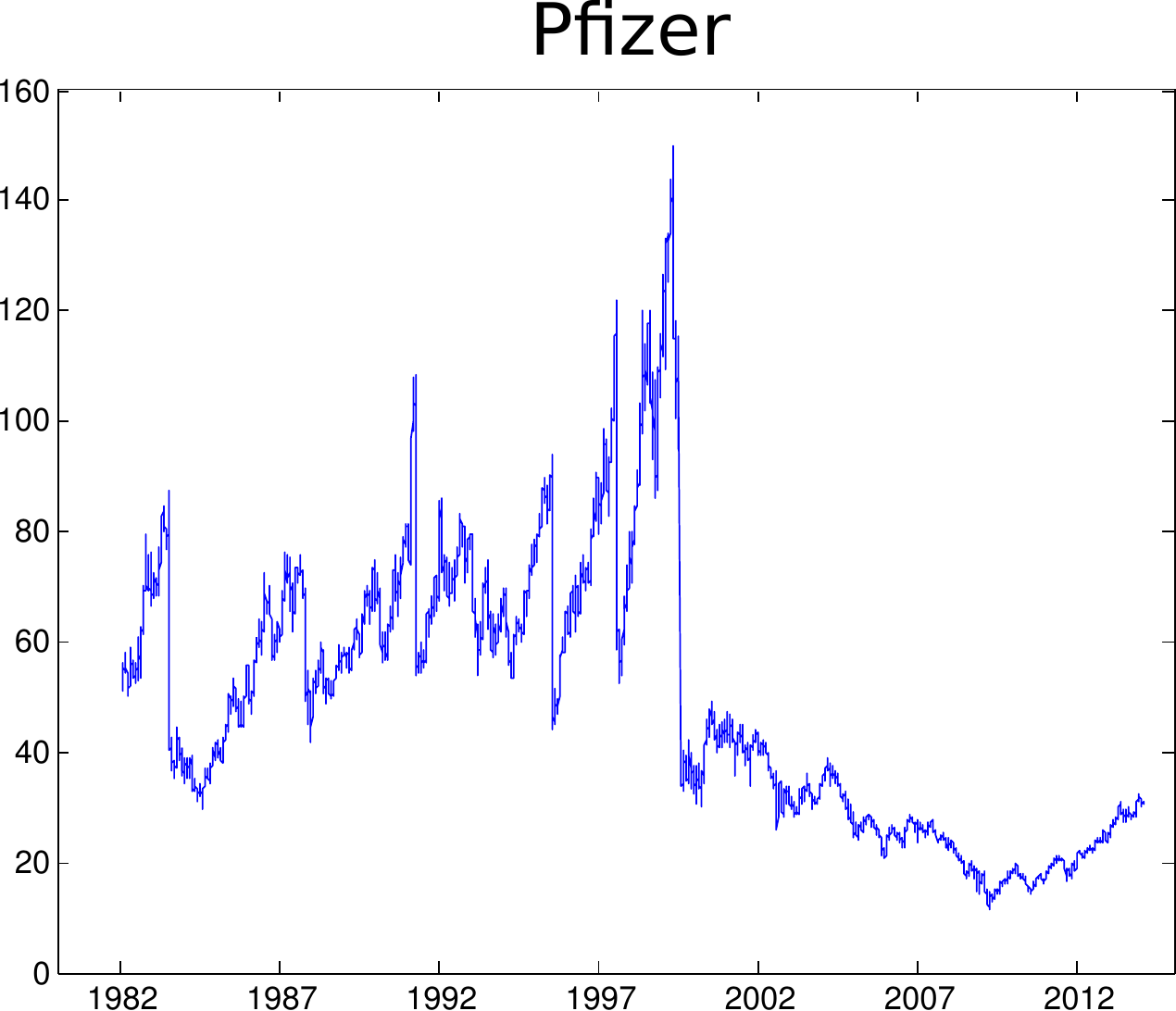}}
    \subfigure{\includegraphics[width=.195\textwidth]{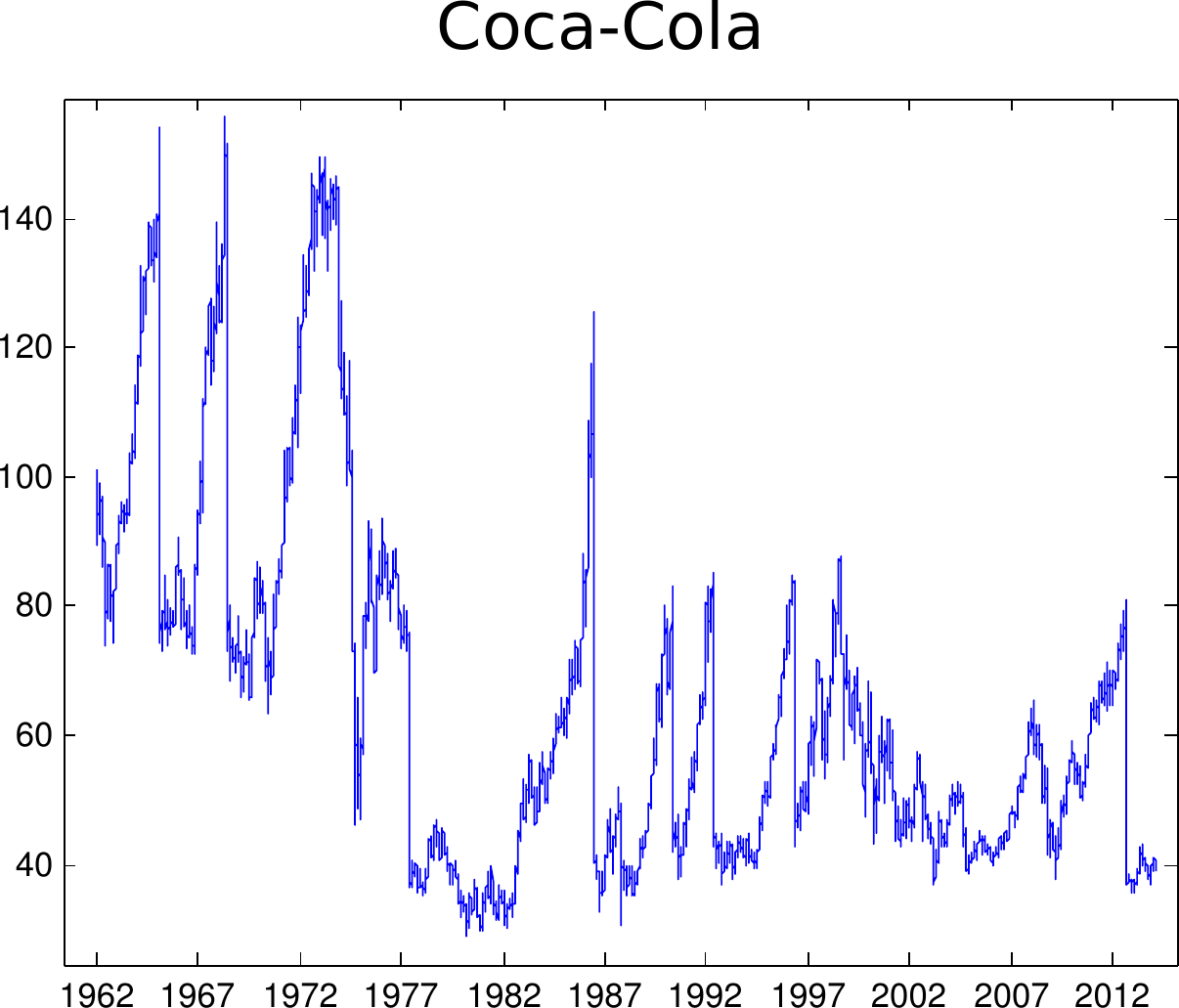}}

  \subfigure{\includegraphics[width=.195\textwidth]{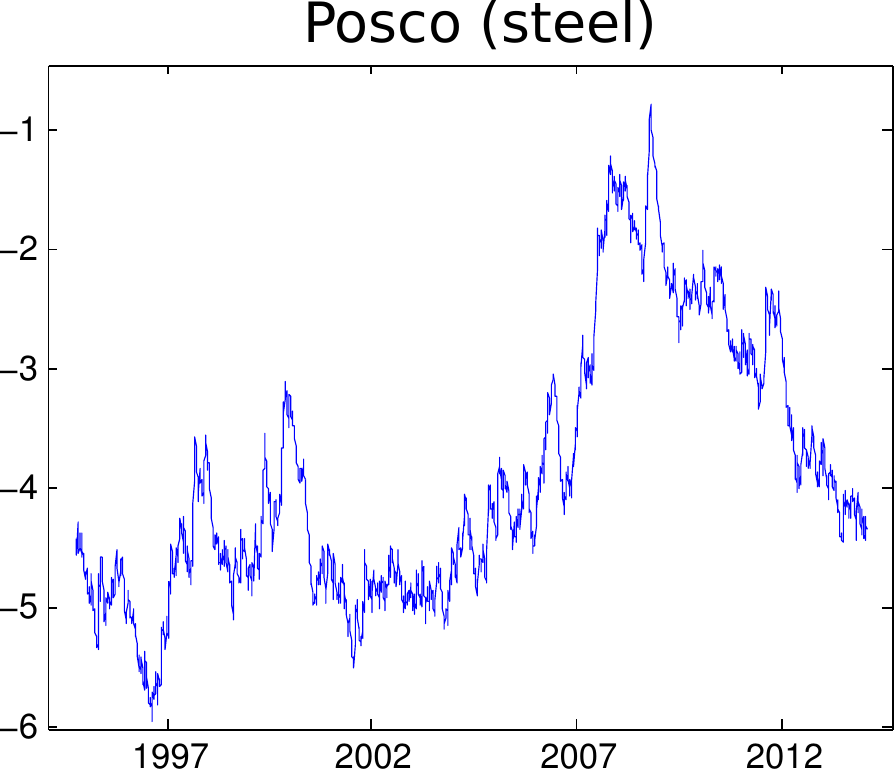}}
  \subfigure{\includegraphics[width=.195\textwidth]{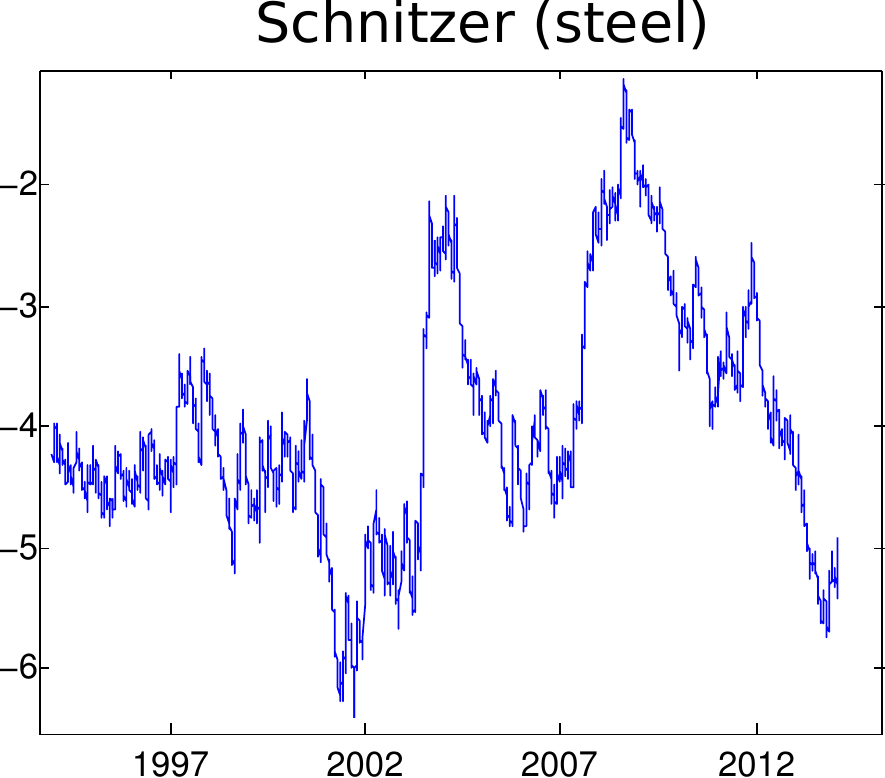}}
 \subfigure{\includegraphics[width=.195\textwidth]{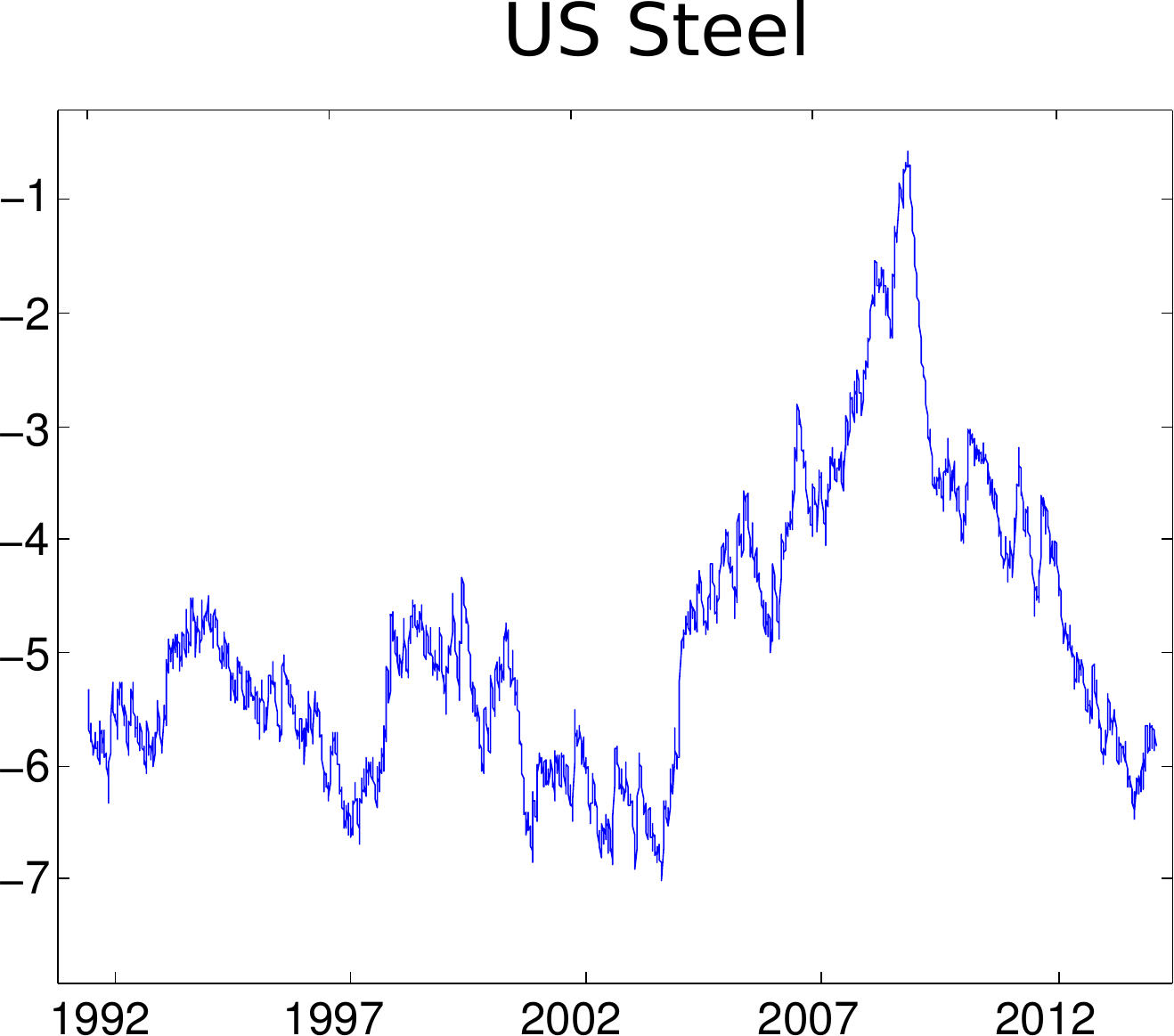}}
   \subfigure{\includegraphics[width=.195\textwidth]{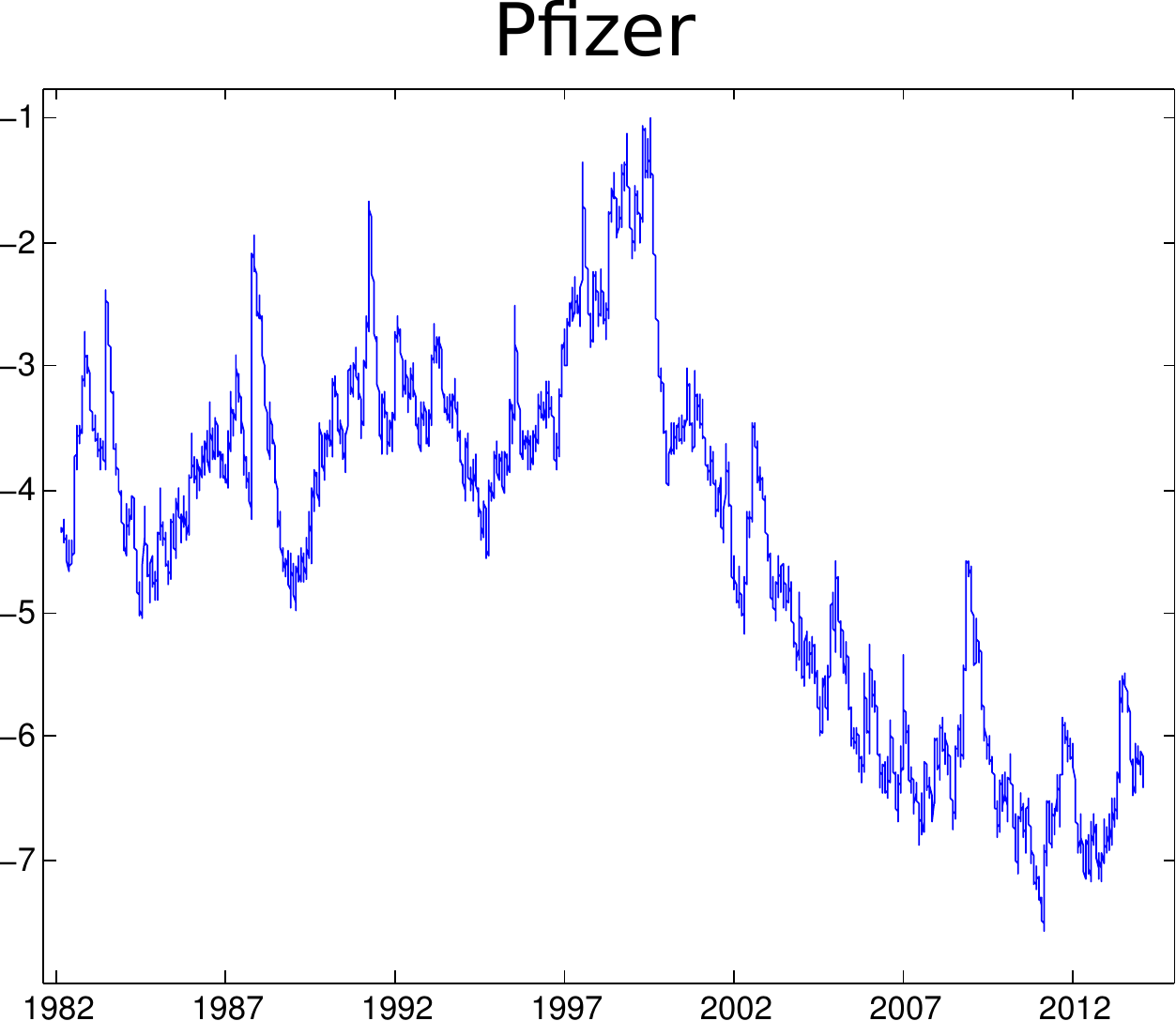}}
 \subfigure{\includegraphics[width=.195\textwidth]{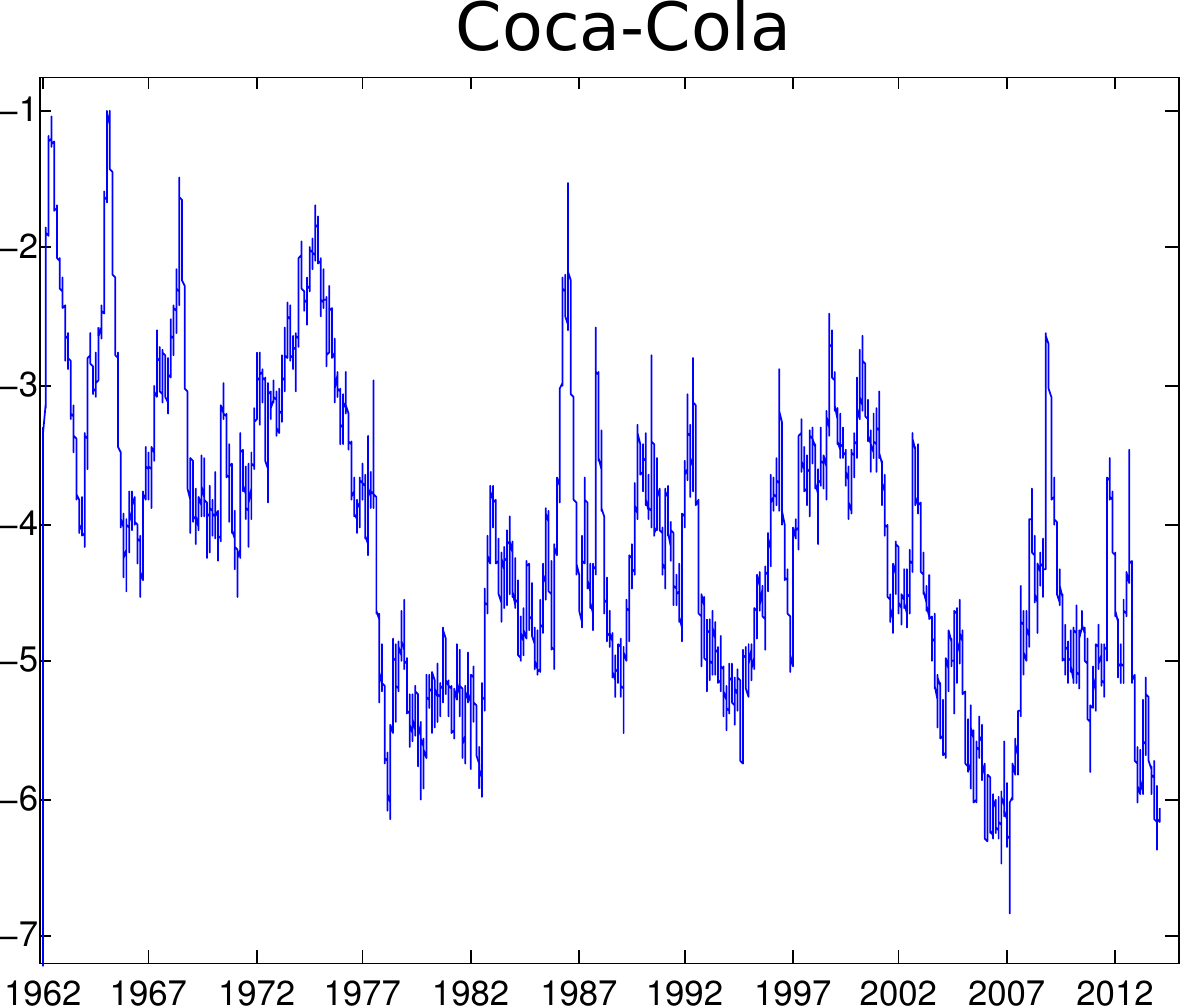}}
 \caption{(Top row) The historical stock prices for three steel companies, one pharmaceutical and one beverage company. (Bottom row) The corresponding log drift Brownian motions ($a_{u_i}\<t\>$) for the respective stocks from the top row. We see that the three steel companies shared high volatility during the period of the 2008 financial crises, but companies in other areas such as the pharmaceutical and beverage industry were not similarly affected.}\label{fig.steel}

\end{figure*}

We also consider the stock volatility across different market sectors. In Figure \ref{fig.steel} we show the historical stock prices for five companies along the top row and their respective $a_{u_i}\<t\>$ below them on the second row. Three of the companies are in the steel market, while the other two are from different markets (pharmaceutical and beverage). We again see that the learned Brownian motion captures a similar volatility for the steel companies. In each case, there is significant volatility associated with the 2008 financial crisis due to the decrease in new construction. This volatility is not found with the pharmaceutical or beverage companies. However, we do see a major spike in volatility for Coca-Cola around 1985, which was a result of their unsuccessful ``New Coke'' experiment. These observations are confirmed by the respective stock prices. However, we note that the level of volatility is not only associated with large changes in stock value. In the Pfizer example, we see that the 
volatility is consistently high for the first half of its stock life, and then decreases significantly for the second half, which is due to the significantly different levels of volatility before and after the year 2000.

\section{Conclusion}
We have presented a collaborative Kalman filter for dynamic collaborative filtering. The model uses the matrix factorization approach, which embeds users and objects in a latent space. However, unlike most related models in which these locations are fixed in time, we allowed them to drift according to a Brownian motion. This allows for dynamic evolution of, e.g., user preference. We built on this basic model by presenting a method for learning a time-evolving drift parameter for the state space vectors using a geometric Brownian motion. 

Many other applications could potentially benefit from this framework. For example, we could predict results of batter/pitcher events over time in baseball or winners of sporting competitions. Such a dynamic model would clearly be an advantage since player or team performance can experience streaks and slumps. Another potential use is in automated tutoring systems, where a student is asked a sequence of questions (e.g., math questions or language flashcards). In a collaborative filtering environment we would wish to estimate the probability that a student answers a question correctly. Clearly as the student learns these probabilities will evolve. Having estimates for such a probability would allow for intelligent tutoring, where questions are selected that are not too easy or too hard and are able to help with the learning process.

\small

\end{document}